\documentclass[smallextended,referee,envcountsect,]{svjour3}
\smartqed

\usepackage{mathtools}
\usepackage{amsmath}

\def\argmin{\textnormal{arg} \min}

\newcommand{\inner}[2]{\langle#1, #2\rangle}

\newcommand{\Br}{\mathbb{R}}
\newcommand{\ones}{\textbf{1}}

\def\RR{\mathbb{R}}

\def\Ss{\mathcal{S}}

\newcommand{\va}{\mathbf{a}}
\newcommand{\vb}{\mathbf{b}}
\newcommand{\vc}{\mathbf{c}}

\newcommand{\vp}{\mathbf{p}}
\newcommand{\vq}{\mathbf{q}}
\newcommand{\vr}{\mathbf{r}}
\newcommand{\vs}{\mathbf{s}}

\newcommand{\vu}{\mathbf{u}}

\newcommand{\vx}{\mathbf{x}}

\newcommand{\vA}{\mathbf{A}}
\newcommand{\vB}{\mathbf{B}}
\newcommand{\vC}{\mathbf{C}}

\newcommand{\vW}{\mathbf{W}}
\newcommand{\vX}{\mathbf{X}}
\newcommand{\vY}{\mathbf{Y}}

\usepackage{graphicx}
\usepackage{bbm}
\usepackage{amsfonts}
\usepackage{amsmath}
\usepackage{amssymb,bbm}

\usepackage{color}
\usepackage{algorithm}
\usepackage{comment}
\usepackage[utf8]{inputenc}
\usepackage[thinc]{esdiff}

\usepackage{thmtools}
\usepackage{thm-restate}

\usepackage{url}
\usepackage[colorlinks=True,linkcolor=magenta,citecolor=black,urlcolor=blue,pagebackref=true]{hyperref}

\usepackage{tabularx}%

\usepackage{enumitem}
\usepackage{booktabs}
\usepackage{multirow}
\usepackage{makecell}
\usepackage{pbox}
\usepackage{float}
\journalname{JOTA}

\begin{document}
\titlerunning{Accelerated Algorithm for Sparse Regularized Partial Optimal Transport}
\authorrunning{K. Nguyen et al.}
\title{Accelerated Algorithm for Sparse Regularized Partial Optimal Transport}
\author{
Khoa Nguyen \and Dung T. Nguyen$^{*}$ \and Thong Huynh$^{*}$ \and Hoang-Hiep Nguyen-Mau \and
Anh Nguyen \and
Minh Ngoc Dinh \and
Juho Kannala
}

\institute{
Khoa Nguyen (Corresponding author) \and Juho Kannala \at
Aalto University, Konemiehentie 2, 02150 Espoo, Finland\\
\email{MarkNguyenDS@gmail.com}\\
\email{juho.kannala@aalto.fi}\\
\emph{Present address of Khoa Nguyen:}
Linköping University, 581 83 Linköping, Sweden
\and
Dung T. Nguyen \at
University of Illinois Urbana-Champaign, Champaign, IL, USA\\
\email{dungn2@illinois.edu}
\and
Thong Huynh \at
Helsinki University,
Fabianinkatu 33, 00100 Helsinki, Finland
\\
\email{thong.minh.huynh@helsinki.fi}
\and
Hoang-Hiep Nguyen-Mau \at
VinUniversity, Hanoi, Vietnam\\
\email{nmhhiep@gmail.com}
\and
Anh Nguyen \at
Rochambeau French International School
9600 Forest Rd, 20814 Bethesda, Maryland, USA\\
\email{minhanhnguyen102008@gmail.com}
\and
Minh Ngoc Dinh \at
Dean of Data Science and IT, Millennia Education, Ho Chi Minh City, Vietnam
\\
\email{minh.dinh@maeducation.com}
}
\begingroup
\renewcommand{\thefootnote}{*}
\footnotetext{Dung T. Nguyen and Thong Huynh contributed equally to this work.}
\endgroup
\date{Received: date / Accepted: date}
\maketitle
\begin{abstract}
Partial Optimal Transport (POT) extends the classical optimal transport problem by relaxing the strict mass conservation constraint, enabling its use in a wide range of real-world applications. In many of these settings, sparse transport plans are preferred for their interpretability and computational benefits. While smooth and strongly convex regularizers—such as quadratic or elastic net—have been vastly used in various machine learning applications to induce sparsity and accelerate computation, they have received less algorithmic attention compared to entropic approaches for computational POT. In this paper, we propose a new optimization framework that leverages these regularizers through a penalty-based reformulation, enabling efficient gradient-based updates while preserving the structure of the original problem. Our method accommodates a broad class of regularizers that promote structured and sparse transport plans. Building on this formulation, we design an accelerated first-order algorithm that alternates between smooth updates and simple projection steps. Through empirical benchmarks on color transfer, domain adaptation, and point cloud registration, our approach consistently outperforms established baselines—achieving lower transport cost, higher sparsity, and faster convergence—making it a practical and scalable solution for modern transport problems.
\end{abstract}
\section{Introduction}


Optimal Transport (OT) \cite{Kantorovich-1942-Translocation,Villani-09} computes a minimum-cost coupling between two mass distributions. Moreover, it quantifies similarity between distributions. These properties have made OT widely applicable in modern machine learning \cite{Arjovsky-2017-Wasserstein,Rabin-Image-Segmentation}. However, its high computational complexity remains a major barrier to broader adoption. In addition, the classical OT formulation enforces a strict mass-balance constraint \cite{peyre2019computational,Villani-09}, requiring the source and target distributions to have equal total mass. Since this assumption is rarely satisfied in practice, it limits the applicability of OT in many real-world scenarios.

Partial Optimal Transport (POT) \cite{chapel2020partialoptimaltransportapplications} addresses this limitation by allowing only a prescribed amount of mass to be transported between two possibly unbalanced distributions. This added flexibility makes POT well-suited to applications involving outliers, missing content, or mismatched sample sizes \cite{chapel2020partialoptimaltransportapplications}. As a result, POT has been used in various machine learning tasks, including color transfer \cite{piticolortransfer,poulicolortransfer}, domain adaptation \cite{redkodomainadaptaion}, and dictionary learning \cite{pmlr-v51-rolet16}.

Despite its modeling advantages, POT remains computationally challenging. Compared to standard OT, the feasible set in POT involves more complex structural constraints, which makes the design of scalable solvers more difficult \cite{our_pot}. To address this, several works \cite{cuturi2013sinkhorn,our_pot,bregman_pot} introduce entropic regularization along with acceleration methods. While these techniques improve runtime, they typically yield dense transport plans \cite{blondel2018smoothsparseoptimaltransport,Tran_2025_Sparse_POT,sparseOT_solomon} due to entropic regularization, which are undesirable in applications that favor sparsity, such as color transfer \cite{piticolortransfer}, domain adaptation \cite{Courty-2017-Optimal}, and ecological inference \cite{Boris-eco-inference}. Another line of work reformulates POT as a standard OT problem by augmenting the marginals. For instance,  \cite{chapel2020partialoptimaltransportapplications} converts inequality constraints into equalities using dummy variables, embedding POT into a higher-dimensional OT problem. This increases both the problem size and the maximum value in the cost matrix, leading to greater memory consumption and computational cost. In particular, since entropic OT solvers depend explicitly on cost magnitudes \cite{Dvurechensky-2018-Computational,lin2019efficient,pmlr-v139-guminov21a}, such reformulations from POT to OT pose scalability issues. To address efficiency issues, \cite{our_pot} adapted the APDAGD algorithm with better complexity and practical performance. However, since the framework strictly relies on entropic regularization, it produces dense transport plans, which limit its applicability in scenarios that require sparsity. 

These limitations motivate the use of smooth and strongly convex regularizers that preserve tractability while promoting sparse transport structures. Quadratic regularization is particularly relevant in this context. For balanced OT, \cite{wiesel2024sparsityquadraticallyregularizedoptimal} quantitatively characterized the sparsity-inducing properties of quadratic regularization. Extending this idea to the partial transport setting, Tran et al. \cite{Tran_2025_Sparse_POT} studied Quadratic-regularized Partial Optimal Transport (QPOT) and empirically benchmarked its sparsity against entropically regularized POT across synthetic datasets and applications such as color transfer and domain adaptation. However, their work focuses on establishing and evaluating the QPOT formulation using general-purpose convex optimization solvers, rather than developing a dedicated scalable optimization algorithm. More broadly, to the best of our knowledge, efficient first-order methods for this class of regularized POT problems remain largely unexplored.

To bridge this computational gap, we propose a penalty-based optimization framework for Partial Optimal Transport with smooth and strongly convex regularization. Rather than introducing another regularized POT formulation alone, we focus on enabling its efficient optimization. The proposed framework includes QPOT as a special case while supporting a broader class of smooth and strongly convex regularizers, including quadratic and elastic-net-type regularization. Building on this formulation, we develop a dedicated accelerated first-order solver that combines gradient-based optimization of the penalized objective with simple projections onto the remaining constraints. This provides a scalable approach for computing sparse regularized POT solutions without relying on general-purpose convex optimization solvers.

\noindent\textbf{Contributions:} We summarize our main contributions as follows:
\begin{itemize}
\item We reformulate Regularized Partial Optimal Transport (RPOT) with smooth and strongly convex regularizers—including quadratic and elastic net regularization—into a penalized optimization problem by introducing an exterior quadratic penalty on the marginal constraints. This removes the need for strict feasibility during intermediate optimization while preserving the structure of the original problem.

\item We prove that the solution to the penalized RPOT formulation is equivalent to that of the original problem. Specifically, with properly chosen penalty parameters, the solution to the penalized problem approximates the RPOT solution up to an arbitrarily small error. We also provide explicit bounds on the objective gap and constraint violations, which help facilitate algorithmic development for the POT problem.

\item By leveraging the smoothness and strong convexity of the penalized objective, we design an accelerated first-order algorithm based on Proximal Nesterov’s method, combined with the rounding strategy from \cite{our_pot} to restore exact feasibility. We show that the algorithm achieves an $\varepsilon$-approximate solution to the original POT problem within $\mathcal{O}\!\left(\frac{n^{7/8}}{\varepsilon}\,\ln\!\left(\frac{n}{\varepsilon}\right)\right)$ iterations with $\mathcal{O}(n^2 \log(n))$ per-iteration cost, while maintaining sparsity and scalability.

\item We validate our framework on various applications: color transfer, domain adaptation, and partial point cloud registration against the Splitting Conic Solver (SCS), and APDAGD solver from \cite{our_pot}. Experimental results demonstrate competitive transport cost, high sparsity, and strong feasibility, with favorable performance compared to existing scalable POT solvers.
\end{itemize}

\label{Introduction}
\section{Preliminaries}
\subsection{Optimal Transport and Partial Optimal Transport}
Before introducing Optimal Transport and its variants, we first present some notation. 

Let 
$ \RR^n_+ := \{\vx \in \RR^n : x_i \geq 0 \ \forall i\} $ 
denote the set of nonnegative vectors, and 
$ \RR^{n\times n}_+ := \{\vX \in \RR^{n\times n} : X_{ij} \geq 0 \ \forall i,j\} $ be the set of nonnegative matrices. Furthermore, for a matrix $\vX \in \RR^{n\times n}$, we denote its following matrix operation and norm as:
\begin{itemize}
    \item $\ell$2 norm of matrix $\vX$: $\|\vX\|_2 = \left(\sum_{i=1}^n \sum_{j=1}^n X_{ij}^2\right)^{1/2}$
    \item infinity norm of matrix $\vX$: $\|\vX\|_\infty = \max_{1 \leq i,j \leq n} |X_{ij}|$
    \item inner product of two matrices, given $\vA, \vB \in \RR^{n \times n}$: $$\langle \vA,\vB \rangle = \sum_{i=1}^n \sum_{j=1}^n A_{ij}B_{ij} = \operatorname{tr}(\vA^\top \vB) $$
    \item element-wise multiplication of two matrices given $\vA, \vB \in \RR^{n \times n}$: $$(\vA \odot \vB)_{ij} = A_{ij}B_{ij} $$
\end{itemize}

This notation will be used throughout the remainder of this section.

\textbf{Optimal Transport (OT):} For two probability distributions with nonnegative and equal total mass $\vr \in \RR^n_+$ and $\vc \in \RR^n_+$, we consider the problem of transporting mass between them at minimum cost. Let $\vC$ be a given cost matrix, and let $\vX$ denote a transportation plan. Following \cite{Kantorovich-1942-Translocation}, the OT problem can be formulated as a linear program:
\begin{align}
\label{OT}
&\mathbf{OT}(\vr, \vc) 
= \min_{\vX \in \mathcal{M}(\vr, \vc)} \ \langle \vC, \vX \rangle, \\
&\mathcal{M}(\vr, \vc) 
:= \{ \vX \in \RR^{n \times n}_+ : \vX \ones_n = \vr, \ \vX^\top \ones_n = \vc \}, \nonumber
\end{align}
where $\mathcal{M}(\vr, \vc)$ denotes the set of feasible transportation plans with prescribed marginals $\vr$ and $\vc$. 

\textbf{Partial Optimal Transport (POT):} The strict requirement of equal total mass in OT can be relaxed by introducing a parameter $s$ representing the total amount of mass to be transported. Consider two discrete distributions $\vr, \vc \in \RR^n_+$ with possibly different total masses. The transported mass $s$ satisfies $0 \leq s \leq \min\{\|\vr\|_1, \|\vc\|_1\}$ \cite{chapel2020partialoptimaltransportapplications,nhatho-mmpot}. The POT problem is then formulated as
\begin{align}
\label{eq:pot_formulation}
&\mathbf{POT}(\vr, \vc, s) = \min_{\vX \in \mathcal{U}(\vr, \vc, s)} \ \langle \vC, \vX \rangle, \\
&\mathcal{U}(\vr, \vc, s) = \big\{ \vX \in \RR_+^{n \times n} : \vX \ones_n \leq \vr,\ \vX^\top \ones_n \leq \vc,\ \ones_n^\top \vX \ones_n = s \big\}, \nonumber
\end{align}
where $\vC \in \RR^{n \times n}_+$ is a cost matrix and $\mathcal{U}(\vr,\vc,s)$ is the feasible set of admissible transport plans. We denote by $\vX^* \in \argmin_{\vX \in \mathcal{U}(\vr, \vc, s)} \langle \vC, \vX \rangle$ an optimal transportation plan of the POT problem.  


Building on this notation, we further introduce the definition of Slater's condition for the POT problem.  

\begin{definition}[Slater's condition for the POT problem]
\label{def:slater-pot}
Given two mass vectors $\vr, \vc \in \RR^n_+$ and a transported mass 
$0 < s < \min\{\|\vr\|_1, \|\vc\|_1\}$, there exists a slack constant $\zeta > 0$ and a strictly feasible
transport plan $\vX^\zeta \in \RR^{n \times n}_+$ such that:
\begin{align}
\vX^\zeta \ones_n \leq \vr - \zeta \ones_n, \qquad
(\vX^\zeta)^\top \ones_n \leq \vc - \zeta \ones_n, \qquad
\ones_n^\top \vX^\zeta \ones_n = s.
\end{align}
\end{definition}

Similar to prior constrained optimization settings, we assume that Slater’s condition holds with slack value $\zeta$ for our upcoming formulation.

\begin{comment}
\begin{definition}[$\varepsilon$-approximation]
\label{eps-approx}
Let $\mathbf{P}_1$ and $\mathbf{P}_2$ denote the optimal values of two optimization problems. For $\varepsilon>0$, we say that $\mathbf{P}_2$ is an $\varepsilon$-approx-\\imation of $\mathbf{P}_1$ if
\begin{equation}
\mid\mathbf{P}_2 - \mathbf{P}_1\mid \leq \varepsilon.
\end{equation}
\end{definition}
\end{comment}

\begin{definition}[$\varepsilon$-convergence]
\label{eps-convergence}
For $\varepsilon > 0$, $\vX \in \RR_+^{n \times n}$ is called an $\varepsilon$-convergent transportation plan of $\mathbf{POT}(\vr, \vc, s)$ if
\begin{equation*}
    \vX \ones_n - \varepsilon \ones_n \leq \vr,
    \qquad
    \vX^\top \ones_n - \varepsilon \ones_n \leq \vc,
    \qquad
    \ones_n^\top \vX \ones_n = s.
\end{equation*}
\end{definition}

\subsection{Regularizers in Partial Optimal Transport}
Regularizers have recently been introduced in OT and POT problems to accelerate the framework (entropic regularizer \cite{cuturi2013sinkhorn}) or encourage structured properties such as sparsity in the optimal transport plan (quadratic regularizer \cite{Tran_2025_Sparse_POT}). Building on these prior works, we define the general Regularized Partial Optimal Transport (RPOT) problem as

\begin{align}
\label{rpot_formulation}
\mathbf{RPOT}_\eta(\vr, \vc, s) &= \min_{\vX \in \mathcal{U}(\vr, \vc, s)} \ \Big\{ f_\eta(\vX) \triangleq \langle \vC, \vX \rangle + \eta R(\vX) \Big\}, \\
\mathcal{U}(\vr, \vc, s) &= \{ \vX \in \RR_+^{n \times n} : \vX \ones_n \leq \vr,\ \vX^\top \ones_n \leq \vc,\ \ones_n^\top \vX \ones_n = s \}, \nonumber
\end{align}
where $\eta > 0$ is a regularization parameter and $R(\vX)$ is a general $\mu -$strongly convex and $L-$smooth regularizer and $R(\vX) \geq 0 \;\forall \vX \in \RR^{n \times n}_+$. Furthermore, we assign \\$\vX^\eta \in \argmin_{\vX \in \mathcal{U}(\vr, \vc, s)} f_\eta(\vX)$ as the optimal transportation plan of the RPOT problem.

\begin{lemma}[Existence of a uniform upper bound]
\label{lma_bound_regularizer}
There exists a finite constant \(U_R>0\) such that
$$
R(\mathbf{X})\leq U_R, \quad \forall \mathbf{X}\in \mathcal U(\vr,\vc,s).
$$
\end{lemma}

\noindent{\it Proof } 
We first show that the feasible set $\mathcal{U}(\vr,\vc,s)$ is compact.
All constraints defining $\mathcal{U}(\vr,\vc,s)$,
namely non-negativity, linear inequalities, and a linear equality,
are linear and hence continuous. Therefore, $\mathcal{U}(\vr,\vc,s)$ is a closed subset of $\mathbb{R}_{+}^{n\times n}$.

Moreover, for any $\mathbf{X}\in\mathcal{U}(\vr,\vc,s)$, since $\mathbf{X}\geq 0$ and
$\mathbf 1^\top \mathbf{X} \mathbf 1 = s$,
it follows that $0\leq X_{ij}\leq s$ for all $i,j$.
Hence, $\mathcal{U}(\vr,\vc,s)$ is bounded.
Being closed and bounded in a finite-dimensional Euclidean space,
$\mathcal{U}(\vr,\vc,s)$ is compact.

Since $R$ is $L-$smooth, and $\mathcal{U}(\vr,\vc,s)$ is compact,
the Weierstrass Extreme Value Theorem \cite{rudin1976principles} guarantees the existence of
$\mathbf{X}_R \in\mathcal{U}(\vr,\vc,s)$ such that
\begin{equation*}
    R(\mathbf{X}_R)=\max_{\mathbf{X}\in \mathcal{U}(\vr,\vc,s)} R(\mathbf{X})
\end{equation*}
Choosing $U_R \geq R(\mathbf{X}_R)$, we obtain a finite upper bound $U_R$ for $R(\vX_R)$. \qed

We list a few regularizers satisfying the boundedness condition in the previous lemma, together with their smoothness parameter $L$, parameter for strong convexity $\mu$, and their upper bound $U_R$.

\begin{enumerate}
\item \emph{Squared $\ell_2$ norm regularizer:}
For $R(\mathbf{X})=\frac{1}{2}\|\mathbf{X}\|^2_2$ and $\mathbf{X}\in\mathcal{U}(\vr,\vc,s)$, we have:
$$\nabla R(\vX) = \vX, \;\, \nabla^2 R(\vX) = \mathbf{I}$$ \\
Then, the smoothness and strong convexity parameters are 
$L=\mu=1$. \\
Furthermore, for any $\vX \in \Ss:= \left\{ \vX \in \mathbb{R}^{n \times n}_+,\; \mathbf 1^\top X \mathbf 1 = s \right\}$: $$\|\vX\|_2^2 = \sum_{i,j} X_{ij}^2
\leq \left( \sum_{i,j} X_{ij} \right)^2
= s^2 = U_R$$



\item \emph{Weighted Element-wise Elastic Net regularizer:} For $\eta_2 > 0$, $\vW_1 \geq 0$ being a matrix of heterogeneous regularization weight for the $\ell_1$ term, and $\vX \in\mathcal{U}(\vr,\vc,s)$, we have: 
\begin{equation*}
\label{elastic-net-reg}
R(X) =\frac{\eta_2}{2}\|\vX\|_2^2 +  \|\mathbf{W}_{1}\odot\vX\|_{1}
= \sum_{i,j}\left(\frac{\eta_2}{2}X_{ij}^2+W_{1, ij}X_{ij}\right)
\end{equation*}
since $|X_{ij}| = X_{ij}$ for $\vX \in \mathcal{U}(\vr, \vc, s)$. Therefore, for all $i,j$, we obtain
$$
\nabla R(X)_{ij} = \eta_2 X_{ij} + \vW_1, \qquad \nabla^2 R(X)_{ij} = \eta_2 I_{mn} $$ where $I_{mn}$ denotes the identity operator on $\mathbb R^{m\times n}$ under vectorization.
\\
Considering the gradient and Hessian of $R$, we obtain: $L = \eta_2, \,\mu = \eta_2$
Finally, for any $X\in\mathcal U$,
$$ R(X) = \frac{\eta_2}{2}\sum_{i,j} X_{ij}^2 + \sum_{i,j}W_{1, ij} X_{ij} \;\leq\; \frac{\eta_2}{2}s^2 + \max_{i,j}\vW_1 s = U_R $$
In the case of uniform $W_{ij}$, the weighted element-wise Elastic Net regularizer becomes the conventional element-wise Elastic Net regularizer.

\begin{comment}
\item \emph{Element-wise Elastic Net regularizer:} For $
R(X)\;=\;\frac{\eta_2}{2}\|\vX\|_2^2\;+\;\eta_1 \|\vX\|_{1}$, let $\eta_1 \geq 0,\eta_2 > 0$ and $\vX \in\mathcal{U}(\vr,\vc,s)$, we have: 
\begin{equation*}
\label{elastic-net-reg}
R(X) =\frac{\eta_2}{2}\|\vX\|_2^2 + \eta_1 \|\vX\|_{1}
= \sum_{i,j}\left(\frac{\eta_2}{2}X_{ij}^2+\eta_1X_{ij}\right)
\end{equation*}
since $|X_{ij}| = X_{ij}$ for $\vX \in \mathcal{U}(\vr, \vc, s)$. Therefore, for all $i,j$, we obtain
$$
\nabla R(X)_{ij} = \eta_2 X_{ij} + \eta_1, \qquad \nabla^2 R(X)_{ij} = \eta_2 I_{mn} $$ where $I_{mn}$ denotes the identity operator on $\mathbb R^{m\times n}$ under vectorization.
\\
Considering the gradient and Hessian of $R$, we obtain: $L = \eta_2, \,\mu = \eta_2$

Finally, for any $X\in\mathcal U$,
$$ R(X) = \frac{\eta_2}{2}\sum_{i,j} X_{ij}^2 + \eta_1 \sum_{i,j} X_{ij} \;\leq\; \frac{\eta_2}{2}s^2 + \eta_1 s = U_R $$
\end{comment}

With the uniform upper bound in Lemma \ref{lma_bound_regularizer} introduced, our generalized RPOT has the following upper bound.
\begin{lemma}
\label{problem_size_lma}
For any $\vX \in \mathcal{U}(\vr, \vc, s)$, we have the bound on problem size for \eqref{rpot_formulation}:
    \begin{align}
        f_\eta(\vX) \leq s \|\vC\|_\infty  + \eta U_R
    \end{align}
where $U_R$ is a generic upper bound of $R(\vX)$
\end{lemma}

{\it Proof } 
Given that: $f_\eta(\vX) \triangleq \inner{\vC}{\vX} + \eta R(\vX)$ where:
\begin{itemize}
    \item $\vC \in \mathbb{R}_+^{n \times n}$, $\vr, \vc \in \mathbb{R}_+^n$, $ s \in \mathbb{R}_+$
    \item $\vX \in \mathcal{U}(\vr, \vc, s)= \{ \vX \in \RR_{+}^{n \times n}: \vX \ones_n \leq \vr, \vX^\top \ones_n \leq \vc, \ones_n^\top \vX \ones_n = s \}$
\end{itemize} 
We denote the upper bound of our smooth and strongly convex regularizer $R(\vX)$ as $U_R$.
\begin{comment}
The regularization term can be re-written as:
\begin{equation*}
\|\mathbf{X}\|_2^2 = \sum_{i=1}^{n} \sum_{j=1}^{n} x_{ij}^2 \leq \sum_{i=1}^{n} \sum_{j=1}^{n} x_{ij}^2 + 2 \sum_{\substack{i=1 \\ (i, j) \neq (k, l)}}^{n} \sum_{j=1}^{n} x_{ij} x_{kl} = \left( \sum_{i=1}^{n} \sum_{j=1}^{n} x_{ij} \right)^2 = \vs^2. \quad \text{(*)}
\end{equation*}
\end{comment}

Now, consider that: $\|\vC\|_\infty = \max_{1 \leq i,j \leq n}C_{ij}$ and $\vC \in \mathbb{R}^{n \times n}_+$. As every element in $\vC$ is nonnegative, this inequality holds: $C_{ij} \leq \|\vC\|_\infty$. Hence, the inner product $\inner{\vC}{\vX}$ can be bounded by:
\begin{equation*}
    \inner{\vC}{\vX} = \sum^n_{i=1}\sum^n_{j=1}C_{ij}X_{ij} \leq \sum^n_{i=1}\sum^n_{j=1}\|\vC\|_\infty X_{ij} = \|\vC\|_\infty\sum^n_{i=1}\sum^n_{j=1}X_{ij} = s\|\vC\|_\infty
\end{equation*}
$f_\eta(\vX)$ hence is bounded by:
\begin{equation*}
    f_\eta(\vX) \triangleq \inner{\vC}{\vX} + \eta R(\vX) \leq s \|\vC\|_\infty  + \eta U_R
\end{equation*}\qed

\end{enumerate}



\section{Penalty Method for RPOT}
Despite the introduction of the regularizer term that promotes sparsity and improves numerical stability, solving RPOT still fundamentally relies on explicit feasibility constraints to define valid transport plans. Enforcing these constraints can be computationally demanding during optimization. To address this limitation, we turn to the framework of exterior penalty functions in \cite{penalty_mazal1}, which provides an approach to transforming constrained optimization problems into unconstrained ones. 

The key idea of the exterior penalty function is to replace hard feasibility constraints with soft penalties that impose an additional cost whenever a constraint is violated. This not only eases the optimization process but also ensures $\varepsilon$-approximation to the solution of the original constrained problem under suitable penalty parameters.  
Formally, consider the generic problem
\begin{align*}
    &\min f(\vx), \quad \text{s.t. } g_i(\vx) \geq 0, \; i=1,\ldots,m,
\end{align*}
where $f(\vx)$ is the objective function and $g_i(\vx), \;i \in \left[1, \ldots, m\right]$ are the constraint functions. 
Under the exterior penalty method, the constrained optimization problem can be reformulated as
\begin{align*}
F(\vx,\alpha) \;\triangleq\; f(\vx) + P(\vx,\alpha)
= f(\vx) + \alpha \sum_{i=1}^m \left[\min\{0, g_i(\vx)\}\right]^2, 
\qquad \alpha > 0,
\end{align*}
where $\alpha$ is a \emph{penalty parameter} that controls the severity of constraint violations. Then, let $\mathcal{T} := \{\vx : g_i(\vx) \ge 0,\ i=1,\dots,m\}$ denote the feasible set of the original problem.
The penalty function $P(\vx,\alpha)$ satisfies:
\begin{align*}
P(\vx,\alpha) &= 0,  \forall\, \vx \in \mathcal{T},\\
P(\vx,\alpha) &> 0, \forall\, \vx \notin \mathcal{T},\\
\lim_{\alpha \to \infty} P(\vx,\alpha) &= \infty, \forall\, \vx \notin \mathcal{T}.
\end{align*}

\subsection{Penalty Function}
In the context of RPOT, not all feasibility constraints pose the same level of difficulty. The nonnegativity constraint $\vX \geq 0$ and the mass constraint $\ones_n^\top \vX \ones_n = s$ are relatively straightforward to enforce. By contrast, the marginal inequality constraints $\vX \ones_n \leq \vr$ and $\vX^\top \ones_n \leq \vc$ are more challenging to handle directly. To alleviate this, with $\alpha \geq 0$, we introduce a quadratic exterior penalty function of the form:
\begin{align}
\label{penalty_func1}
&P(\vX,\alpha) = \alpha \sum_{i=1}^n \Big( \min\{0, r_i - (\vX \ones_n)_i\}^2 + \min\{0, c_i - (\vX^\top \ones_n)_i\}^2 \Big)
\end{align}
Denote $\mathcal{T} = \{\vX \in \RR^{n \times n}: \vX \ones_n \leq \vr, \ \vX^\top \ones_n \leq \vc\}$ as the feasible set associated with the marginal inequality constraints. For the remaining simple constraints, let $\Ss = \{\vX \in \RR^{n \times n}_+:\ones_n^\top \vX \ones_n = s  \}$ be their feasible domain. Hence, for any $\alpha > 0$, the \textbf{penalized regularized partial optimal transport problem (P-RPOT)} is formally defined as
\begin{align}
\label{pqpot_objective}
&\mathbf{P\text{-}RPOT}_{\eta,\alpha}(\vr,\vc,s) 
= \min_{\vX \in \Ss} \ \Big\{ F_\eta(\vX,\alpha) \triangleq f_\eta(\vX) + P(\vX,\alpha) \Big\}, \\
&= \min_{\vX \in \Ss} \ \Big\{f_\eta(\vX) + \alpha \sum_{i=1}^n \Big( \min\{0, r_i - (\vX \ones_n)_i\}^2 + \min\{0, c_i - (\vX^\top \ones_n)_i\}^2 \Big)\Big\}
\label{pqpot_objective1}
\end{align}

According to Lemma \ref{lma_converging_penalty}, the function $P(\vX,\alpha)$ satisfies the requirements of an exterior penalty method as in \cite{penalty_mazal1}. That is, \eqref{penalty_func1} is monotonically increasing in $\alpha$ and strictly penalizes any solution $\vX$ that is not in $\mathcal{T}$ and hence encourages the solution to converge to $\vX \in \mathcal{T}$. Therefore, \eqref{penalty_func1} provides a valid relaxation of the original RPOT.

\begin{lemma}
\label{lma_converging_penalty}
For $0 < \alpha_1 < \alpha_2$, we have the following:
\begin{align}
    P(\vX, \alpha_1) \leq P(\vX, \alpha_2).
\end{align}
Moreover, for any $\alpha > 0$, we have:
\begin{align}
P(\vX, \alpha) &= 0, \forall \vX \in \mathcal{T} \\
P(\vX, \alpha) &> 0, \forall \vX \not\in \mathcal{T}, \\
\lim_{\alpha \to \infty} P(\vX, \alpha) &= \infty,  \forall \vX \not\in \mathcal{T}.
\end{align}
 \end{lemma}
{\it Proof } 
Given $\vX \in \mathcal{T} = \{\vX\in \Br^{n \times n}: \vX \ones_n \leq \vr, \vX^\top \ones_n \leq \vc\}$, we can rewrite the first two constraints as:
\begin{align*}
\mathbf{X}\mathbf{1}_n \leq \mathbf{r}
&\Leftrightarrow \mathbf{r}_i - (\mathbf{X} \mathbf{1}_n)_i \geq 0 \;\Leftrightarrow \min(\mathbf{r}_i - (\mathbf{X} \mathbf{1}_n)_i, 0) = 0 \quad \forall i \quad (\star)
\\
\mathbf{X}^\top \mathbf{1}_n \leq \mathbf{c} 
&\Leftrightarrow \mathbf{c}_i - (\mathbf{X}^\top \mathbf{1}_n)_i \geq 0 \Leftrightarrow \min(\mathbf{c}_i - (\mathbf{X}^\top \mathbf{1}_n)_i, 0) = 0 \quad \forall i \quad (\star \star)
\end{align*}
From $(\star), (\star\star)$:
\begin{align*}
P(\mathbf{X}, \alpha)
&= \alpha \sum_{i=1}^{n} \left( \min \{ 0, \mathbf{r}_i - (\mathbf{X} \mathbf{1}_n)_i \}^2 + \min \{ 0, \mathbf{c}_i - (\mathbf{X}^\top \mathbf{1}_n)_i \}^2 \right) = 0
\end{align*}
On the other hand, if $\vX \notin \mathcal{T}$, then there exists $i \in \{1, \cdots, n\}$ such that either $$\mathbf{r}_i - (\mathbf{X} \mathbf{1}_n)_i < 0 \quad \text{or} \quad \mathbf{c}_i - (\mathbf{X}^\top \mathbf{1}_n)_i < 0$$ 
Hence $$\sum_{i=1}^n\left(\min \{ 0, \mathbf{r}_i - (\mathbf{X} \mathbf{1}_n)_i \}^2 + \min \{ 0, \mathbf{c}_i - (\mathbf{X}^\top \mathbf{1}_n)_i \}^2\right) = k > 0$$ 
Hence, $P(\mathbf{X}, \alpha) = \alpha k > 0$

\noindent Then:
$$P(\mathbf{X}, \alpha_1) \leq P(\mathbf{X}, \alpha_2) \quad \text{for $k > 0$, $\alpha_2 > \alpha_1 > 0$}$$
As $P(\mathbf{X}, \alpha)$ monotonically increases with $\alpha$ when $\vX \notin \mathcal{T}$, the following holds $$\lim_{\alpha\to\infty}P(\vX, \alpha) =\lim_{\alpha\to\infty}\alpha k = \infty$$\qed

Given that $P(\vX,\alpha)$ grows with $\alpha$ for all $\vX \notin \mathcal{T}$, Theorem \ref{thm_equiv_champ} establishes that, for sufficiently large $\alpha \to \infty$, minimizing the penalized problem is equivalent to minimizing $\mathbf{RPOT}_\eta(\vr,\vc,s)$. 

\begin{theorem}
\label{thm_equiv_champ}
    We have:
    \begin{align}
    \label{equiv_champ}
          \textbf{RPOT}_{\eta}(\vr, \vc, s) =  \lim_{\alpha \to \infty} \textbf{P-RPOT}_{\eta, \alpha}(\vr, \vc, s)
    \end{align}
\end{theorem}

\noindent{\it Proof } 
Fix $\eta>0$. The sets $\Ss$ and $\mathcal{U}(\vr,\vc,s)$ are
nonempty and compact, and the objectives are continuous. Hence, 
$ \textbf{RPOT}_{\eta}(\vr, \vc, s)$ and $\textbf{P-RPOT}_{\eta, \alpha}(\vr, \vc, s)$ admit minimizers, denoted by
$X^\eta$ and $X^{\eta,\alpha}$, respectively.

Since $X^\eta$ is feasible for $ \textbf{RPOT}_{\eta}(\vr, \vc, s)$, we have
$P(X^\eta,\alpha)=0$. Optimality of $X^{\eta,\alpha}$ and
nonnegativity of the objective terms therefore give
\begin{align}
\nonumber
0
&\le \textbf{P-RPOT}_{\eta,\alpha}(\vr,\vc,s) = f_\eta(X^{\eta,\alpha})+P(X^{\eta,\alpha},\alpha) \\
&\le f_\eta(X^\eta)+P(X^\eta,\alpha) = \textbf{RPOT}_{\eta}(\vr,\vc,s).\label{inter_proof_yeah1}
\end{align}

Moreover, the penalized optimal value is nondecreasing in $\alpha$,
because $P(X,\alpha)$ is nondecreasing in $\alpha$ for every
$X\in \Ss$. Thus its limit as $\alpha\to\infty$ exists and is bounded
above by $\textbf{RPOT}_{\eta}(\vr,\vc,s)$.

To establish the reverse inequality, observe that $f_\eta\ge0$ on
$S$ and $P(X,\alpha)=\alpha P(X,1)$ imply
\begin{align}
0\le P(X^{\eta,\alpha},1)
=\frac{P(X^{\eta,\alpha},\alpha)}{\alpha}
\le\frac{\textbf{RPOT}_{\eta}(\vr,\vc,s)}{\alpha}
\overset{\alpha\to \infty}{\longrightarrow}0. \label{tozero_step}
\end{align}
Since $\Ss$ is compact, by the Bolzano–Weierstrass Theorem, there exists a convergent subsequence  
\begin{align*}
X^{\eta,\alpha_k}\longrightarrow\widehat X\in \Ss,
\end{align*}
such that $\alpha_k\to\infty$.
By \eqref{tozero_step} and the continuity of $P(\,\cdot\,,1)$, we have $P(\widehat X,1)=0$.
Hence, $\widehat X$ satisfies the marginal inequalities and, since
$\widehat X\in S$, belongs to $\mathcal{U}(\vr,\vc,s)$.
Consequently, feasibility of $\widehat X$ and continuity of
$f_\eta$ give
\[
\begin{aligned}
\textbf{RPOT}_{\eta}(\vr,\vc,s)
&\le f_\eta(\widehat X) = \lim_{k\to\infty}f_\eta(X^{\eta,\alpha_k}) \\
&\le \lim_{k\to\infty}
     \textbf{P-RPOT}_{\eta,\alpha_k}(\vr,\vc,s) \\
&\overset{\eqref{inter_proof_yeah1}}{\le} \textbf{RPOT}_{\eta}(\vr,\vc,s).
\end{aligned}
\]
All inequalities are therefore equalities, proving the claim.
\qed

Here, let $\vX^{\eta,\alpha}$ denote the optimal solution to $\mathbf{P\text{-}RPOT}_{\eta,\alpha}(\vr,\vc,s)$. As we have already established such equivalence between the two problems, we show in the next subsection that, for sufficiently large but finite $\alpha$, the solution $\vX^{\eta,\alpha}$ can also be made to approximate the true RPOT solution $\vX^\eta$ with an error of $\varepsilon'$.


\subsection{Convergence to the true solution of RPOT}
As discussed above, Theorem \ref{thm_equiv_champ} shows that, for sufficiently large $\alpha$, the solution of \textbf{P-RPOT} converges to that of \textbf{RPOT}. Building on this result, Theorem \ref{thm_strongbabe1} derives a lower bound on $\alpha$ that guarantees a small objective gap between $\vX^{\eta,\alpha}$ and $\vX^\eta$. Furthermore, assuming that \textbf{RPOT} satisfies Slater's condition, we show that $\vX^{\eta,\alpha}$ also attains a controlled constraint violation, with both up to a sufficiently small error.

\begin{theorem}
\label{thm_strongbabe1}
    Under Slater's condition (Definition~\ref{def:slater-pot}) for \textbf{RPOT}, for any $\varepsilon' > 0$ and $$\alpha > \frac{(\sqrt{2n}+1)(s\|\vC\|_\infty + \eta U_R)^2}{2\zeta^2\varepsilon'} = \mathcal{O}\big(\frac{\sqrt{n}}{\varepsilon'}\big)$$, let $\vX^{\eta, \alpha} = \argmin_{\vX \in \mathcal{S}}F_\eta(\vX, \alpha)$ be the solution to $\textbf{P-RPOT}_{\eta, \alpha}(\vr, \vc, s)$, we have:
    \begin{align}
    \label{equiv_champ_full}
          \textbf{RPOT}_{\eta}(\vr, \vc, s) =  \textbf{P-RPOT}_{\eta, \alpha}(\vr, \vc, s) \pm \varepsilon'.
    \end{align}
    Furthermore, the constraint violation is bounded as
    \begin{align}
        \vX^{\eta, \alpha} \ones_n - \tau_\alpha \ones_n \leq \vr \label{rolex1}\\
    \vX^{\eta, \alpha \top }\ones_n - \tau_\alpha \ones_n \leq \vc \label{rolex2}
    \end{align}
    where $\tau_\alpha < \frac{\zeta \varepsilon'}{s\|\vC\|_\infty + \eta U_R} = \mathcal{O}(\varepsilon')$.
\end{theorem}

{\it Proof } 
Under Slater's condition and by Theorem 1 in \cite{penalty_mazal1} (restated in Appendix~\ref{app:truemper}), the quadratic exterior penalty method admits the following guarantee: for any tolerance $\tau > 0$, setting the penalty parameter as
\begin{equation}
\label{eq:tau_to_alpha}
\alpha = \frac{\alpha_0 h}{\tau},
\end{equation}
with $h = \frac{\sqrt{2n} + 1}{2}$, ensures that the solution $\vX^{\eta, \alpha}$ is a $\tau$-convergent transportation plan in the sense of Definition \ref{eps-convergence}. Here, $h$ comes from the fact that $P(\vX, \alpha)$ acts on $m = 2n$ inequality constraints, namely the marginal inequalities $\vX \ones_n \leq \vr$ and $\vX^\top \ones_n \leq \vc$.

It remains to specify the penalty threshold $\alpha_0$. Let $\vX^0$ denote an optimal solution of the original constrained problem, and let $\bar{\vX}$ be any strictly feasible point (i.e., $g_i(\bar{\vX}) > 0\, \forall i$). Translating inequality (8) of \cite{penalty_mazal1} from its maximization setting to our minimization setting (equivalently, applying it to $\max\{-f_\eta\}$), if $z$ is a lower bound on $f_\eta(\vX^0)$, then $\alpha_0$ may be chosen to satisfy
\begin{equation}
    \alpha_0 \geq \frac{f_\eta(\bar{\vX}) - z}{\min_i g_i(\bar{\vX})}.
\end{equation}
Since $f_\eta(\vX) \geq 0$ for all feasible $\vX$, in particular $f_\eta(\vX^0) \geq 0$, we may take $z = 0$, which reduces the bound to
\begin{equation}
    \alpha_0 \geq \frac{f_\eta(\bar{\vX})}{\min_i g_i(\bar{\vX})}. \label{eq:alpha0_start}
\end{equation}
To obtain an explicit threshold independent of the particular choice of $\bar{\vX}$, we upper-bound the right-hand side of \eqref{eq:alpha0_start}. By Lemma \ref{problem_size_lma}, the numerator is bounded above at every feasible point, and in particular at $\bar{\vX}$:
\begin{equation}
    f_\eta(\bar{\vX}) \leq s \|\vC\|_\infty + \eta U_R. \label{eq:f_eta_bound}
\end{equation}
For the denominator, Slater's condition guarantees a constant $\zeta > 0$ such that
\begin{equation}
    \min_i g_i(\bar{\vX}) \geq \zeta \Leftrightarrow \frac{1}{\min_i g_i(\bar{\vX})} \leq \frac{1}{\zeta}, \label{eq:slater_bound}
\end{equation}
Combining \eqref{eq:f_eta_bound} and \eqref{eq:slater_bound} with \eqref{eq:alpha0_start}, it suffices to take
\begin{equation}
    \alpha_0 = \frac{s \|\vC\|_\infty + \eta U_R}{\zeta}. \label{eq:alpha0_bound}
\end{equation}

With $\alpha_0$ pinned down by \eqref{eq:alpha0_bound}, Theorem 1 in \cite{penalty_mazal1} applies with the threshold $\alpha_0 + \delta$ for any $\delta > 0$. Since $\vX^{\eta, \alpha}$ does not depend on $\delta$, letting $\delta \to 0$ in \eqref{eq:tau_to_alpha} gives the feasibility tolerance induced by a given $\alpha$:
\begin{equation}
\label{eq:tau_alpha_def}
    \tau_\alpha := \frac{\alpha_0 h}{\alpha} = \frac{(\sqrt{2n}+1)(s\|\vC\|_\infty + \eta U_R)}{2\zeta\alpha},
\end{equation}
under which $\vX^{\eta,\alpha}$ is $\tau_\alpha$-convergent in the sense of Definition \ref{eps-convergence}, which is exactly \eqref{rolex1} and \eqref{rolex2}. The simplified form of $\tau_\alpha$ stated in the theorem follows once we derive the required lower bound on $\alpha$ from the objective-gap analysis below; we defer this simplification until then.

It remains to derive the objective gap \eqref{equiv_champ_full} from the $\tau_\alpha$-level constraint violation above. We show the two directions of \eqref{equiv_champ_full} separately.

The first direction is immediate. Since $\vX^\eta$ is feasible for \textbf{RPOT}, we have $P(\vX^\eta, \alpha) = 0$, hence
\begin{equation}
\label{eq:gap_direction1}
    \textbf{P-RPOT}_{\eta,\alpha} = F_\eta(\vX^{\eta,\alpha}, \alpha) \leq F_\eta(\vX^\eta, \alpha) = f_\eta(\vX^\eta) = \textbf{RPOT}_\eta.
\end{equation}

For the second direction, we adapt the sensitivity argument outlined in \cite[p.~605]{penalty_mazal1} to our convex setting. For any $t \geq 0$, define the $t$-relaxed feasible set and value function
\begin{equation*}
    \mathcal{U}_t := \{\vX \in \RR_+^{n\times n} : \vX\ones_n \leq \vr + t\ones_n,\; \vX^\top \ones_n \leq \vc + t\ones_n,\; \ones_n^\top \vX \ones_n = s\},
\end{equation*}
\begin{equation*}
    v(t) := \min_{\vX \in \mathcal{U}_t} f_\eta(\vX),
\end{equation*}
so that $v(0) = \textbf{RPOT}_\eta$. The $\tau_\alpha$-convergence established above implies $\vX^{\eta,\alpha} \in \mathcal{U}_{\tau_\alpha}$, hence
\begin{equation}
    v(\tau_\alpha) \leq f_\eta(\vX^{\eta,\alpha}) \leq F_\eta(\vX^{\eta,\alpha}, \alpha) = \textbf{P-RPOT}_{\eta,\alpha}. \label{eq:value_dom}
\end{equation}
What remains is to control the drop $v(0) - v(\tau_\alpha)$. Under Slater's condition, strong duality holds for \textbf{RPOT}, so there exist optimal Lagrange multipliers $\vu^0 \in \RR_+^{2n}$ associated with the $2n$ marginal inequalities. By the multiplier bound following inequality (9) of \cite{penalty_mazal1}, these multipliers satisfy $\|\vu^0\|_1 \leq \alpha_0$. Since $v(\cdot)$ is convex and non-increasing, in light of \cite[\S5.6.2]{boyd_convex_2004}), the sensitivity bound for convex programs under Slater's condition gives
\begin{equation}
\label{eq:value_drop}
    v(0) - v(\tau_\alpha) \leq \|\vu^0\|_1 \cdot \tau_\alpha \leq \alpha_0 \cdot \tau_\alpha = \frac{\alpha_0^2 h}{\alpha} = \frac{(\sqrt{2n}+1)(s\|\vC\|_\infty + \eta U_R)^2}{2\zeta^2\alpha}.
\end{equation}

To bound the objective gap by $\varepsilon'$, we require the right-hand side of \eqref{eq:value_drop} to be at most $\varepsilon'$, i.e.
\begin{align*}
    \label{eq:alpha_derived}
    &\frac{(\sqrt{2n}+1)(s\|\vC\|_\infty + \eta U_R)^2}{2\zeta^2\alpha} \leq \varepsilon'
    \quad \\\Leftrightarrow\quad
    &\alpha \geq \frac{(\sqrt{2n}+1)(s\|\vC\|_\infty + \eta U_R)^2}{2\zeta^2 \varepsilon'} = \mathcal{O}\!\left(\frac{\sqrt{n}}{\varepsilon'}\right),
\end{align*}
which is exactly the lower bound on $\alpha$ stated in the theorem. Substituting this $\alpha$ back into \eqref{eq:tau_alpha_def} also yields the simplified form of $\tau_\alpha$ deferred earlier:
\begin{equation}
\label{eq:tau_alpha_simplified}
    \tau_\alpha < \frac{\zeta \varepsilon'}{s\|\vC\|_\infty + \eta U_R} = \mathcal{O}(\varepsilon').
\end{equation}
Combining \eqref{eq:value_dom} with \eqref{eq:value_drop} bounded by $\varepsilon'$ gives $\textbf{RPOT}_\eta = v(0) \leq v(\tau_\alpha) + \varepsilon' \leq \textbf{P-RPOT}_{\eta,\alpha} + \varepsilon'$, which together with \eqref{eq:gap_direction1} proves \eqref{equiv_champ_full}. \qed

\section{Complexity Analysis of Approximating RPOT}
In this section, we will formally introduce our main algorithm for approximating P-RPOT, together with its supporting auxiliary algorithms. Before doing so, we first recall and define the necessary quantities.

\begin{table}[h!]
\caption{Notation and Definitions of Main Variables}
\centering
\begin{tabular}{cl}
\hline
\textbf{Notation} & \textbf{Description} \\
\hline
$\vX^*$ & Solution of original POT: 
$
\displaystyle \vX^* \in \arg\min_{\vX \in \mathcal{U}} \langle \vC, \vX \rangle
$ \\[10pt]

$\vX^\eta$ & Solution of RPOT: 
$
\displaystyle \vX^\eta \in \arg\min_{\vX \in \mathcal{U}} f_\eta(\vX)
$ \\[10pt]

$\vX^{\eta, \alpha}$ & Solution of P-RPOT: $
\displaystyle \vX^{\eta, \alpha} \in \arg\min_{\vX \in \mathcal{S}} F_{\eta}(\vX, \alpha)
$ \\[10pt]

$\bar{\vX}$ & Final output of ROUND-POT algorithm \\[8pt]

$\vX^k$ & Output at iteration $k$ of PNAG-POT solver \\[3pt]
\hline
\end{tabular}
\label{tab:solutions}
\end{table}

Table \ref{tab:solutions} summarizes the notation for the solutions of the original POT, the RPOT, and the P-RPOT problems, as well as the solutions produced by our proposed Nesterov’s P-RPOT solver (Algorithm \ref{alg:PNAG}) and its rounded version obtained through the ROUND-POT algorithm (Algorithm \ref{alg:RoundPOT}). Table \ref{tab:errors} then introduces the corresponding error terms that describe the relationships among these solutions, quantifying how solving P-RPOT approximates the original POT and RPOT objectives.

\begin{table}[h!]
\centering
\caption{Notation and Definitions of Error Terms}
\begin{tabular}{cl}
\hline
\textbf{Notation} & \textbf{Description} \\
\hline
$\varepsilon$ & \parbox[t]{0.78\linewidth}{Objective gap between the rounded  and original POT solution:
$$
    \langle \vC, \bar{\vX} \rangle - \langle \vC, \vX^* \rangle \leq \varepsilon
$$} \\[6pt]

$\varepsilon'$ & \parbox[t]{0.78\linewidth}{Difference between P-RPOT and RPOT objectives:\\[3pt]
$$
0 \leq f_\eta(\vX^{\eta}) - F_\eta(\vX^{\eta, \alpha}, \alpha) \leq \varepsilon'
$$
} \\[6pt]

$\varepsilon_1$ & \parbox[t]{0.78\linewidth}{Nesterov solver error:\\[3pt]
$$
0 \leq F_{\eta}(\vX^k, \alpha) - F_{\eta}(\vX^{\eta, \alpha}, \alpha) \leq \varepsilon_1
$$
} \\[3pt]
\hline
\end{tabular}
\label{tab:errors}
\end{table}

With these quantities established, Section \ref{sec5.2} discusses the smoothness and convexity properties of P-RPOT, which form the theoretical foundation for our proposed approach in Section \ref{sec5.3}. Finally, Section \ref{sec5.4} provides a complexity analysis of obtaining the P-RPOT solution that is $\varepsilon$-approximation to RPOT.

\subsection{Smoothness and Convexity of P-RPOT} \label{sec5.2}

Recall that Theorems \ref{thm_equiv_champ} and \ref{thm_strongbabe1} establish the equivalence between \\$\textbf{RPOT}_{\eta}(\vr, \vc, s)$ and $\textbf{P-RPOT}_{\eta, \alpha}(\vr, \vc, s)$ up to a controlled error when \(\alpha\) is sufficiently large. Hence, the \textbf{RPOT} problem can be solved through the strongly convex and smooth \textbf{P-RPOT} formulation of \eqref{pqpot_objective}:

\begin{align}
\label{pqpot_objective_new}
     &\textbf{P-RPOT}_{\eta, \alpha}(\vr, \vc, s) = \min_{ \vX \in \Ss}\Big\{ F_\eta(\vX,\alpha) \Big\} \notag \\
     &\text{where:}\notag\\
     &F_\eta(\vX,\alpha) \triangleq f_\eta(\vX) + \alpha \sum_{i=1}^n \Big( \min\{0, r_i - (\vX \ones_n)_i\}^2 + \min\{0, c_i - (\vX^\top \ones_n)_i\}^2 \Big) \notag \\
     &= \langle \mathbf{C}, \mathbf{X} \rangle  + \eta R(\vX) + \alpha \sum_{i=1}^n\Big( \min\{0, \vr_i - (\vX \ones_n)_i \}^2 + \min\{0, \vc_i - (\vX^\top \ones_n)_i \}^2 \Big)
\end{align}

where the marginal inequality constraints are penalized, and the remaining constraints reduce to the scaled simplex
\begin{equation*}
    \Ss=\{\vX \in \RR_+^{n\times n}:\ones_n^\top\vX\ones_n=s\}
\end{equation*}

In Lemma \ref{lma_convex_smooth}, we further establish the strong convexity and smoothness parameters of \eqref{pqpot_objective_new}, which are crucial to the convergence properties of our proposed gradient-based algorithm.

\begin{comment}
From Theorem \ref{thm_strongbabe1}, we now know that solving the original problem $\textbf{RPOT}_{\eta}(\vr, \vc, s)$ is ultimately solving the problem \eqref{pqpot_objective} $\textbf{P-RPOT}_{\eta, \alpha}(\vr, \vc, s)$ where \eqref{pqpot_objective} corresponds to strongly-convex and smooth optimization with simple constraints: the box constraint $\vX \in \RR_{+}^{n \times n}$ and the $\ell_1$ ball constraint $ \ones_n^\top \vX \ones_n = s$. Recall the formulation of \eqref{pqpot_objective} which could be further expanded as:
\begin{align}
\label{pqpot_objective_new}
     &\textbf{P-RPOT}_{\eta, \alpha}(\vr, \vc, s) = \min_{ \vX \in \Ss}\Big\{ F_\eta(\vX,\alpha) \triangleq f_\eta(\vX) + P(\vX,\alpha) \Big\} \notag \\
     &= f_\eta(\vX) + \alpha \sum_{i=1}^n \Big( \min\{0, r_i - (\vX \ones_n)_i\}^2 + \min\{0, c_i - (\vX^\top \ones_n)_i\}^2 \Big) \notag \\
     &= \langle \mathbf{C}, \mathbf{X} \rangle  + \eta R(\vX) + \alpha \sum_{i=1}^n\Big( \min\{0, \vr_i - (\vX \ones_n)_i \}^2 + \min\{0, \vc_i - (\vX^\top \ones_n)_i \}^2 \Big) 
\end{align}
we could derive the strong convexity and smoothness parameter of \eqref{pqpot_objective_new} in Lemma \ref{lma_convex_smooth}. These parameters are key to our proposed gradient-based algorithm.
\end{comment}
\begin{lemma}
\label{lma_convex_smooth}
The function \eqref{pqpot_objective_new} is an $\eta\mu$-strongly convex and $\beta$-smooth function where $\beta = \eta L +  4 \alpha n = O(n \alpha)$ \\

\end{lemma}
{\it Proof } 
The term $\langle \mathbf{C}, \mathbf{X} \rangle$ is linear and convex. Moreover, the
function $u \mapsto \min(0,u)^2$ is convex, and $(\mathbf{X}\ones_n)_i$ and
$(\mathbf{X}^\top\ones_n)_j$ are linear in $\vX$. Hence, the penalty term $P(\vX, \alpha)$
is convex.

Finally, since $R(\vX)$ is $\mu$-strongly convex, $\eta R(\vX)$ is
$(\eta \mu)$-strongly convex. As the sum of a convex function and an
$(\eta\mu)$-strongly convex function is $(\eta\mu)$-strongly convex, we conclude that
$F_{\eta}(\mathbf{X}, \alpha)$ is $(\eta\mu)$-strongly convex on $\Ss$.

To prove the smoothness of $F_{\eta}(\mathbf{X}, \alpha)$, we consider its gradient.
For each $(i,j)$, we have
\begin{equation}
\frac{\partial F_{\eta}(\mathbf{X,\alpha})}{\partial X_{ij}}
= C_{ij} + \eta \frac{\partial R(\mathbf{X})}{\partial X_{ij}}
+ \frac{\partial P(\vX, \alpha)}{\partial X_{ij}} ,
\label{eq:grad_penalty}
\end{equation}
where, differentiating the two squared violations in $P(\vX,\alpha)$,
\begin{align}
\frac{\partial P(\vX, \alpha)}{\partial X_{ij}}
&= - 2\alpha \Big( r_i - \sum_{k=1}^n X_{ik} \Big)
   \mathbb{I}\Big( r_i - \sum_{k=1}^n X_{ik} < 0 \Big) \notag\\
&\quad - 2\alpha \Big( c_j - \sum_{k=1}^n X_{kj} \Big)
   \mathbb{I}\Big( c_j - \sum_{k=1}^n X_{kj} < 0 \Big).
\label{eq:penalty_grad}
\end{align}

Let $\vX, \vY \in \mathcal{S}$. Taking the difference of \eqref{eq:penalty_grad} at
$\vX$ and at $\vY$,
\begin{align}
    &\frac{\partial P(\vX, \alpha)}{\partial X_{ij}}
     - \frac{\partial P(\vY, \alpha)}{\partial X_{ij}} \notag\\
    =\;&\Big[-2\alpha \Big( r_i - \sum_{k=1}^n X_{ik} \Big)
    \mathbb{I}\Big( r_i - \sum_{k=1}^n X_{ik} < 0 \Big) \notag\\
    &\quad +2\alpha \Big( r_i - \sum_{k=1}^n Y_{ik} \Big)
    \mathbb{I}\Big( r_i - \sum_{k=1}^n Y_{ik} < 0 \Big)\Big] \notag\\
    &+\Big[- 2\alpha \Big( c_j - \sum_{k=1}^n X_{kj} \Big)
    \mathbb{I}\Big( c_j - \sum_{k=1}^n X_{kj} < 0 \Big) \notag \\
    &\quad + 2\alpha \Big( c_j - \sum_{k=1}^n Y_{kj} \Big)
    \mathbb{I}\Big( c_j - \sum_{k=1}^n Y_{kj} < 0 \Big)\Big].
    \label{eq:gradient_diff}
\end{align}

Consider the first bracket in \eqref{eq:gradient_diff}:
\begin{align*}
&2\alpha \Big[ -\Big( r_i - \sum_{k=1}^n X_{ik} \Big)\mathbb{I}\Big( r_i - \sum_{k=1}^n X_{ik} < 0 \Big) + \Big( r_i - \sum_{k=1}^n Y_{ik} \Big)\Big]
    \mathbb{I}\Big( r_i - \sum_{k=1}^n Y_{ik} < 0 \Big) \Big]\\
    &=\begin{cases}
    0, & \text{if } \sum_{k=1}^n X_{ik} \leq r_i,\; \sum_{k=1}^n Y_{ik} \leq r_i, \\[4pt]
    2\alpha \Big( \sum_{k=1}^n X_{ik} - \sum_{k=1}^n Y_{ik} \Big),
      & \text{if } \sum_{k=1}^n X_{ik} > r_i,\; \sum_{k=1}^n Y_{ik} > r_i, \\[4pt]
    2\alpha \Big( \sum_{k=1}^n X_{ik} - r_i \Big),
      & \text{if } \sum_{k=1}^n Y_{ik} \leq r_i < \sum_{k=1}^n X_{ik}, \\[4pt]
    2\alpha \Big( r_i - \sum_{k=1}^n Y_{ik} \Big),
      & \text{if } \sum_{k=1}^n X_{ik} \leq r_i < \sum_{k=1}^n Y_{ik}.
    \end{cases}
\end{align*}
The last two cases give
\begin{align*}
    0 &< 2\alpha \Big( \sum_{k=1}^n X_{ik} - r_i \Big)
      \leq 2\alpha \Big( \sum_{k=1}^n X_{ik} - \sum_{k=1}^n Y_{ik} \Big), \\
    0 &> 2\alpha \Big( r_i - \sum_{k=1}^n Y_{ik} \Big)
      \geq 2\alpha \Big( \sum_{k=1}^n X_{ik} - \sum_{k=1}^n Y_{ik} \Big).
\end{align*}
Hence, in all cases, the absolute value of the first bracket is bounded by
\begin{align*}
    &\Big| -2\alpha \Big( r_i - \sum_{k=1}^n X_{ik} \Big)
    \mathbb{I}\Big( r_i - \sum_{k=1}^n X_{ik} < 0 \Big)
    +2\alpha \Big( r_i - \sum_{k=1}^n Y_{ik} \Big)
    \mathbb{I}\Big( r_i - \sum_{k=1}^n Y_{ik} < 0 \Big) \Big| \\
    &\leq\; 2\alpha \Big| \sum_{k=1}^n X_{ik} - \sum_{k=1}^n Y_{ik} \Big|
    \;\leq\; 2\alpha \sum_{k=1}^n \big| X_{ik} - Y_{ik} \big|,
\end{align*}
where the first inequality is the bound for the four cases above and the second from the
triangle inequality.

The same logic applied for the second bracket. Hence, we can bound \eqref{eq:gradient_diff}
\begin{align}
    & \big\| \nabla P(\vX, \alpha) - \nabla P(\vY, \alpha) \big\| \notag\\
    =\;&\sqrt{\sum^n_{i=1}\sum^n_{j=1}
      \Big[\frac{\partial P(\vX, \alpha)}{\partial X_{ij}}
         - \frac{\partial P(\vY, \alpha)}{\partial X_{ij}}\Big]^2} \notag \\
    \leq\;&\sqrt{\sum^n_{i=1}\sum^n_{j=1}\Big[2\alpha\sum^n_{k=1}\big|X_{ik}-Y_{ik}\big|
      + 2\alpha\sum^n_{k=1}\big|X_{kj}-Y_{kj}\big|\Big]^2} \notag \\
    \leq\;&\sqrt{2n\sum^n_{i=1}\Big(2\alpha\sum^n_{k=1}\big|X_{ik}-Y_{ik}\big|\Big)^2
      + 2n\sum^n_{j=1}\Big(2\alpha\sum^n_{k=1}\big|X_{kj}-Y_{kj}\big|\Big)^2}  \notag \\
    \leq\;&\sqrt{2n\sum^n_{i=1}4\alpha^2n\sum^n_{k=1}\big|X_{ik}-Y_{ik}\big|^2
      + 2n\sum^n_{j=1}4\alpha^2n\sum^n_{k=1}\big|X_{kj}-Y_{kj}\big|^2} \;\text{(Cauchy-Schwarz)}\notag \\
    =\;&\sqrt{2n\big(4\alpha^2n\|\vX - \vY\|^2\big) + 2n\big(4\alpha^2n\|\vX - \vY\|^2\big)}
    \;=\;4\alpha n\|\vX - \vY\| ,
    \label{eq:bound_grad_penalty}
\end{align}
where the second inequality follows from $(p+q)^2 \leq 2p^2 + 2q^2$ applied to each
squared term, the first bracket being independent of $j$ and the second of $i$, so that
the summation over the remaining index repeats each value $n$ times.
Applying \eqref{eq:bound_grad_penalty} to \eqref{eq:grad_penalty}, together with the
$L$-smoothness of $R$ and the triangle inequality, we obtain
\begin{equation*}
\big\|\nabla F_{\eta}(\vX, \alpha) - \nabla F_{\eta}(\vY, \alpha)\big\|
\leq \eta L\|\vX - \vY\| + 4\alpha n\|\vX - \vY\|
= (\eta L + 4\alpha n)\|\vX - \vY\| .
\end{equation*}
Hence, $F_{\eta}(\mathbf{X,\alpha})$ is $(\eta\mu)$-strongly convex and
$(\eta L + 4\alpha n)$-smooth on $\mathcal{S}$.
\qed

\subsection{Gradient Method for P-RPOT} \label{sec5.3}
As aforementioned, solving $\textbf{RPOT}_{\eta}(\vr, \vc, s)$ boils down to minimizing the smooth and strongly convex function \eqref{pqpot_objective_new}. Hence, we base our approach on the Proximal Nesterov's Accelerated Gradient Descent introduced in \cite{nesterov_book}.

On the other hand, since gradient-based algorithms do not take into account the remaining nonnegativity constraint and mass constraint, we use the projection algorithm of \cite{simplex_projection} (Algorithm \ref{alg:simplex_projection} in Appendix \ref{app:alg_projection}) with $\mathcal{O}(n^2\log(n))$ complexity to project our solution to the feasible region $\Ss$. The projection is performed after each step of the gradient iteration.

Finally, to ensure feasibility of the solution, we apply the ROUND-POT rounding procedure (Algorithm \ref{alg:RoundPOT}), which has $\mathcal{O}(n^2)$ complexity, together with its auxiliary Algorithm \ref{alg:EP} from \cite{our_pot}.

The variables defined in Algorithm \ref{alg:RoundPOT},  $\vp \text{ and } \vq \in \RR^n_+$, are the slack variables of the two inequality constraints of the original RPOT problem, such that they account for the violation of these constraints. Thus, we further introduce the following lemma, which shows that when the constraint violation is small, the ROUND-POT algorithm return a feasible solution close to the input $\vX$ .
\begin{lemma}
    \label{slack_bound}
    If the marginal constraint violation of the solution found by PNAG-POT algorithm $\vX^k \in \Ss$ is by a small quantity $\delta$:
    \begin{equation*}
        \sum^n_{i=1}\left[\max\{0, (\vX^k\ones_n)_i -  \vr_i\} + \max\{0,  ((\vX^k)^\top\ones_n)_i - \vc_i\}\right] \leq \delta
    \end{equation*}
    Then, the feasible solution $\bar{\vX}$ output by ROUND-POT is guaranteed to be close to the input solution $\vX^k$ such that:
    \begin{equation*}
    \|\bar{\vX} - \vX^k\|_1 \leq 23\delta
    \end{equation*}
\end{lemma}
{\it Proof } 
Similar to the work of \cite{our_pot}  and as mentioned in Algorithm \ref{alg:RoundPOT}, we define the two slack variables $\vp^k$ and $\vq^k$ as follows:
    \begin{align*}
        p^k_i = \max\{0,  \vr_i - \vX_i^k    \mathbf{1}_n \} \geq 0 \quad\text{and} \quad
        q^k_i =\max\{0,  \vc_i - ((\vX^k)^\top)_i    \mathbf{1}_n \} \geq 0
    \end{align*}

With the introduction of $\vp^k$ and $\vq^k$, we further reformulate our constraints by introducing $\vA$ and $\vb$ using vectorization with $\vx^k = (\operatorname{vec}(\vX^k)^\top, (\vp^k)^\top, (\vq^k)^\top)^\top$, such that the constraint of our optimization problem now has the form:
    \begin{align*}
        &\vA\vx = \vb \quad \text{where } \vA \in \mathbb{R}^{(2n+1)\times(n^2+2n)}, \;\vb \in \mathbb{R}^{2n+1}.\\
        \text{such that } 
           &\vA \vx^k = 
        \begin{pmatrix}
        (\vX^k\ones_n + \vp^k)^\top ,
        (\vX^{k\top}\ones_n + \vq^k)^\top,
        \ones^\top\vX^k\ones
        \end{pmatrix}^\top, 
        \\
        &\vb =
        \begin{pmatrix}
        \vr^\top,
        \vc^\top,
        s
        \end{pmatrix}^\top
    \end{align*}

Recall the definition of $\vA$ and $\vb$ and use the fact that $\ones_n^\top\vX^k\ones_n = s$
\begin{align*}
    &\left\|\vA \vx^k - \vb\right\|_1
    = \left\| \begin{bmatrix}
    \left( (\vX^k \ones_n + \vp^k)^\top - \vr^\top\right),
    \left( ((\vX^k)^\top \ones_n + \vq^k)^\top - \vc^\top \right),
    \left(\ones_n^\top \vX^k \ones_n - s \right)
    \end{bmatrix}\right\|_1 \\ 
    &= \sum^n_{i=1} \left|\vX_i^k \ones_n + p^k_i - \vr_i \right| + \sum^n_{i=1} \left| ((\vX^k)^\top)_i \ones_n + q^k_i - \vc_i \right| + \left|\ones_n^\top\vX^k\ones_n - s \right|  \\
    &= \sum^n_{i=1} \left[ \mid \vX_i^k \ones_n + p^k_i - \vr_i \mid +  \mid ((\vX^k)^\top)_i \ones_n + q^k_i - \vc_i \mid\right]
\end{align*}
Consider the quantity $\vX_i^k \ones_n - \vr_i$, if:
\begin{itemize}
    \item \textbf{Case 1}: $\vX_i^k \ones_n - \vr_i \leq 0 \Rightarrow p^k_i = \max(0, \vr_i - \vX^k_i\ones_n) =  \vr_i - \vX^k_i\ones_n$: \\
    $\mid\vX_i^k \ones_n + p^k_i - \vr_i\mid =\mid \vX_i^k \ones_n + \vr_i - \vX^k_i\ones_n - \vr_i\mid = 0 = -\min\{0, \vr_i - (\vX^k\ones_n)_i\} $
    \\
    \item \textbf{Case 2}: $\vX_i^k \ones_n - \vr_i > 0 \Rightarrow p^k_i = \max(0, \vr_i - \vX^k_i\ones_n) = 0$: \\
    $\mid \vX_i^k \ones_n + p^k_i - \vr_i \mid = \vX_i^k \ones_n - \vr_i = -(\vr_i - \vX_i^k \ones_n) = -\min\{0, \vr_i - (\vX^k\ones_n)_i\} $
        
\end{itemize}
Similarly for the quantity $\mid ((\vX^k)^\top)_i \ones_n + q^k_i  - \vc_i \mid$, we can conclude: given $\vA$ and $\vb$, and let $\vX^k$ be the solution returned at the end of the PNAG-POT algorithm \text{(Algorithm \ref{alg:PNAG})}, we can formally derive the equivalence 
\begin{align}
    \label{eq:slack_equiv}
    \|\vA \vx^k -\vb\|_1 &= \sum^n_{i=1}\left[-\min\{0, \vr_i - (\vX^k\ones_n)_i\} - \min\{0, \vc_i - ((\vX^k)^\top \ones_n)_i\}\right] \notag\\
    &= \sum^n_{i=1}\left[\max\{0,  (\vX^k\ones_n)_i - \vr_i\} + \max\{0,  ((\vX^k)^\top \ones_n)_i - \vc_i\}\right] \leq \delta
\end{align}

Since $\vX^k \geq 0$ and $\vp^k, \vq^k \geq 0$, we have $\vx^k \geq 0$. In view of Theorem 6 from \cite{our_pot} (details in Appendix \ref{app:theorem6ourpot}) and the equivalence we achieved, we reach the final bound:
\begin{equation*}
   \|\bar{\vX} - \vX^k\|_1 \leq  \|\bar{\vx} - \vx^k\|_1 = \|\bar{\vX} - \vX^k\|_1 +  \|\bar{\vp} -\vp^k \|_1 + \|\bar{\vq} - \vq^k\|_1  \leq 23\delta  
\end{equation*}
\qed

\begin{figure}[H]
\centering
\begin{minipage}[t]{0.56\linewidth}
\begin{algorithm}[H]
\caption{Enforcing Procedure EP}
\label{alg:EP}
\begin{minipage}{\linewidth}
\textbf{Input:} marginal $\mathbf{r}$; mass $s$; slack variable $\mathbf{p}$.
\vspace{1mm}

\begin{enumerate}
  \item $\mathbf{p}' = \min\{\mathbf{p}, \mathbf{r}\},\\ 
        \psi = \min\{1,\ (\|\mathbf{r}\|_1 - s)/\|\mathbf{p}'\|_1\}$
  \item $\mathbf{p}'' = \psi \mathbf{p}'$
  \item \textbf{if} $1 > (\|\mathbf{r}\|_1 - s)/\|\mathbf{p}'\|_1$ \textbf{then}
  \item \quad $\bar{\mathbf{p}} = \mathbf{p}''$
  \item \textbf{else}
  \item \quad $i = 0$
  \item \quad \textbf{while} $\|\mathbf{p}''\|_1 \le \|\mathbf{r}\|_1 - s$ \textbf{do}
  \item \qquad $i = i + 1$
  \item \qquad $p''_i = r_i$
  \item \quad \textbf{end while}
  \item \quad $p''_i = p''_i - (\|\mathbf{p}''\|_1 - \|\mathbf{r}\|_1 + s)$
  \item \quad $\bar{\mathbf{p}} = \mathbf{p}''$
  \item \textbf{end if}
\end{enumerate}

\vspace{1mm}
\textbf{Output:} $\bar{\mathbf{p}}$
\end{minipage}
\end{algorithm}
\end{minipage}
\hfill
\begin{minipage}[t]{0.42\linewidth}
\begin{algorithm}[H]
\caption{ROUND-POT}
\label{alg:RoundPOT}
\begin{minipage}{\linewidth}
\textbf{Input:} $\mathbf{X}$, marginals $\mathbf{r}, \mathbf{c}$; mass $s$.
\vspace{1mm}

\begin{enumerate}
  \item Define $\vp = \max\{0, \vr - \vX\ones_n\}$,\\ $\vq = \max\{0, \vc - \vX^\top\ones_n\}$
  \item $\bar{\mathbf{p}} = \operatorname{EP}(\mathbf{r}, s, \mathbf{p})$
  \item $\bar{\mathbf{q}} = \operatorname{EP}(\mathbf{c}, s, \mathbf{q})$
  \item $\mathbf{g} = \min\{\,\mathbf{1},\ (\mathbf{r} - \bar{\mathbf{p}}) \oslash (\mathbf{X}\mathbf{1})\,\}$
  \item $\mathbf{h} = \min\{\,\mathbf{1},\ (\mathbf{c} - \bar{\mathbf{q}}) \oslash (\mathbf{X}^\top\mathbf{1})\,\}$
  \item $\mathbf{X}' = \operatorname{diag}(\mathbf{g})\,\mathbf{X}\,\operatorname{diag}(\mathbf{h})$
  \item $\mathbf{e}_1 = (\mathbf{r} - \bar{\mathbf{p}}) - \mathbf{X}'\mathbf{1}$
  \item $\mathbf{e}_2 = (\mathbf{c} - \bar{\mathbf{q}}) - (\mathbf{X}'^\top\mathbf{1})$
  \item $\bar{\mathbf{X}} = \mathbf{X}' + \dfrac{\mathbf{e}_1 \mathbf{e}_2^\top}{\|\mathbf{e}_1\|_1}$
\end{enumerate}

\vspace{1mm}
\textbf{Output:} $\bar{\mathbf{x}} = (\bar{\mathbf{X}}, \bar{\mathbf{p}}, \bar{\mathbf{q}})$
\end{minipage}
\end{algorithm}
\end{minipage}
\end{figure}

With the ROUND-POT algorithm established, we formally introduce our solver, Algorithm \ref{alg:PNAG}, Proximal Nesterov's Accelerated Gradient algorithm (PNAG-POT).  

\begin{algorithm}[h!]
\caption{Proximal Nesterov's Accelerated Gradient for RPOT (PNAG-POT)}
\label{alg:PNAG}
\begin{minipage}{\linewidth}
\textbf{Require:} Initial value $\mathbf{X}_0 \in \mathbb{R}^{n \times n}$, set momentum term $\mathbf{V}_0 = \mathbf{X}_0$, parameter $\theta \in (0, 1]$
\vspace{1mm}

\textbf{Set:} 
\begin{itemize}
    \item  $i = \{ 0, 1, \ldots, k\}$ be the iteration number
    \item Step size $t_i$ either fixed as $t_i = 1/\beta, \beta=\eta L + 4\alpha n$ or computed via line search.
    \item $\operatorname{prox}$ being the projection algorithm \ref{alg:simplex_projection}
\end{itemize}

\vspace{1mm}
\textbf{Repeat until convergence:}
\begin{enumerate}
  \item Compute $\gamma_i = \dfrac{\theta_{i-1}^2}{t_{i-1}}$
  \item Compute $\theta_i$ as the positive solution to the equation:
  $$
  \frac{\theta_i^2}{t_i} = (1 - \theta_i)\gamma_i + \eta\mu\theta_i
  $$
  \item Set intermediate point $\mathbf{Y}$ as:
  $$
  \mathbf{Y} = 
  \begin{cases}
  \mathbf{X}^0, & \text{if } i = 0 \\
  \mathbf{X}^i + \dfrac{\theta_i \gamma_i}{\gamma_i + \eta\mu \theta_i} (\mathbf{V}^i - \mathbf{X}^i), & \text{if } i > 0
  \end{cases}
  $$
  \item Update:
  $$
  \mathbf{X}_{i+1} = \operatorname{prox}_{t_i}(\mathbf{Y} - t_i\nabla f(\mathbf{Y}))
  $$
  \item Update momentum term:
  $$
  \mathbf{V}_{i+1} = \mathbf{X}_i + \frac{\mathbf{X}_{i+1} - \mathbf{X}_i}{\theta_i}
  $$
\end{enumerate}
\textbf{End} where $\vX^k$ be the converged solution

$\bar{\vX} = \text{ROUND-POT}(\vX_k)$ 

\vspace{1mm}
\textbf{Return:} $\bar{\vX}$
\end{minipage}
\end{algorithm}

\subsection{Iteration Complexity of Approximating RPOT} \label{sec5.4}

In the following Theorem, we establish the number of iterations for our PNAG-POT algorithm to converge to a low-error solution that is:
\begin{align}
    K_0 = \sqrt{\frac{\beta}{\eta\mu}}\left[\ln\!\left(\frac{64\cdot 47^{2}\,n^{2}\,s^{2}\,\|C\|_{\infty}^{2}}{\varepsilon^{2}}\right) + \ln\!\left(\frac{23}{3}\right)\right] = \mathcal{O}\left(\frac{n^{7/8}}{\varepsilon}\ln\left(\frac{n}{\varepsilon}\right)\right)\label{final_iteration_complexity}
\end{align}
\begin{theorem}
\label{thm_final}
For $\eta = \frac{\varepsilon}{8U_R}$, $\alpha=\frac{(s \|\vC\|_\infty  + \eta s^2)^2(\sqrt{2n}+1)}{2\zeta^2 \varepsilon'} = \mathcal{O}\big(\frac{ n^{0.5}}{\varepsilon'} \big)$, and $k \geq K_0$, the output $\bar{\vX}$ of Algorithm \ref{alg:PNAG} is an $\varepsilon-$ approximate solution to the solution of the original POT problem $\vX^*$ such that:
\begin{equation*}
    \inner{\vC}{\bar{\vX}} \leq \inner{\vC}{\vX^*} + \varepsilon, \quad \bar{\vX}, \vX^* \in \mathcal{U}(\vr, \vc, s)
\end{equation*}
\end{theorem}

{\it Proof } 
Recall from Lemma \ref{lma_convex_smooth}, $F_{\eta}(\vX, \alpha)$ is $\eta \mu$-strongly convex and $\beta$-smooth with $\beta = \eta L + 4n\alpha$.
Now, from the convergence guarantees in \cite[Section 2.2]{nesterov_book} and given any $\varepsilon_1 > 0$, for
\begin{align}
k \geq \sqrt{\kappa} \bigg[ \ln\big( \frac{\eta\mu \| \vX^0 - \vX^{\eta,\alpha}\|^2_2}{2\varepsilon_1}\big) + \ln\big(\frac{23}{3}\big) \bigg],  \label{main_proof1}
\end{align}
where $\kappa = \frac{\beta}{\eta\mu}$ and $\beta = \eta L + 4n\alpha$, the $k^{th}$ iterate $\vX^k$ returned by the PNAG algorithm achieves the objective gap of $\varepsilon_1$ from the optimal objective, i.e.
\begin{align}
0\leq  F_{\eta}(\vX^k, \alpha) - F_{\eta}(\vX^{\eta,\alpha}, \alpha) \leq    \varepsilon_1.   \label{main_proof2}
\end{align}
Furthermore, by the first-order optimality condition and the strong convexity properties proven in Lemma \ref{lma_convex_smooth}, we derive the bound for $\|\vX^k - \vX^{\eta, \alpha}\|_2$, which helps us derive the bound for $\|\vX^k - \vX^{\eta, \alpha}\|_1$. 

\begin{lemma}
\label{lma_transition_bound}
    We have the following bound for any $\vX\in \Ss$:
    \begin{align}
        \|\vX- \vX^{\eta,\alpha}\|_2 \leq \sqrt{2}(\eta\mu)^{-\frac{1}{2}} | F_{\eta}(\vX, \alpha) - F_{\eta}(\vX^{\eta,\alpha}, \alpha)|^{\frac{1}{2}}
    \end{align}
\end{lemma}
{\it Proof }  
Since $\vX^{\eta,\alpha}$ is the optimal solution to $\min_{\vX \in \Ss}  F_{\eta}(\vX, \alpha) $, by the first-order optimality condition, we first have:
\begin{align}
       \langle \nabla   F_{\eta}(\vX^{\eta,\alpha}, \alpha), \vX - \vX^{\eta,\alpha} \rangle \geq 0. \label{proof_lma_transition_bound1}
\end{align}
By Lemma \ref{lma_convex_smooth}, $F_{\eta}(\vX, \alpha)$ is $(\eta\mu)$-strongly convex in $\Ss$. Thus, we have:
\begin{align*}
     F_{\eta}(\vX, \alpha) &\geq  F_{\eta}(\vX^{\eta,\alpha}, \alpha) +      \langle \nabla   F_{\eta}(\vX^{\eta,\alpha}, \alpha), \vX - \vX^{\eta,\alpha} \rangle  + \frac{\eta\mu}{2} \|\vX - \vX^{\eta,\alpha}      \|_2^2\\
     &\overset{\eqref{proof_lma_transition_bound1} }{\geq}  F_{\eta} (\vX^{\eta,\alpha}, \alpha) +  \frac{\eta\mu}{2} \|\vX - \vX^{\eta,\alpha}      \|_2^2,
\end{align*}
which thus implies the required statement $$ \| \vX- \vX^{\eta,\alpha}\|_2 \leq \sqrt{2}(\eta\mu)^{-\frac{1}{2}} | F_{\eta}(\vX, \alpha) - F_{\eta}(\vX^{\eta,\alpha}, \alpha)|^{\frac{1}{2}}$$ \qed

Using Lemma \ref{lma_transition_bound}, we arrive at the bound for $\|\vX^k - \vX^{\eta, \alpha}\|_1$:
\begin{align}
\nonumber
   \|\vX^k -\vX^{\eta,\alpha} \|_1 &\leq n \|\vX^k -\vX^{\eta,\alpha} \|_2 \leq \sqrt{2}n (\eta\mu)^{-\frac{1}{2}}| F_{\eta}(\vX^k, \alpha) - F_{\eta}(\vX^{\eta,\alpha}, \alpha)|^{\frac{1}{2}} \\
   &\overset{\eqref{main_proof2}}{\leq } \sqrt{2}n (\eta\mu)^{-\frac{1}{2}}\varepsilon_1^{\frac{1}{2}}. \label{main_proof_4th_term}
\end{align}

In the following steps, we prove that applying ROUND-POT is non-trivial and helps achieve an $\varepsilon$-approximate solution $\vX^*$ of the original POT problem. With that in mind, we derive the error between the output $\bar{\vX}$ of Algorithm \ref{alg:PNAG} and the original solution $\vX^*$:
\begin{align}
    \nonumber
     &\langle \mathbf{C}, \bar{\vX} \rangle - \langle \mathbf{C}, \vX^* \rangle  \\&=   
     \langle \mathbf{C}, \vX^{\eta, \alpha} - \vX^\eta \rangle  +   
     \langle \mathbf{C}, \vX^\eta - \vX^* \rangle + 
     \langle \mathbf{C}, \bar{\vX}-     \vX^k \rangle + 
     \langle \mathbf{C}, \vX^k - \vX^{\eta, \alpha}\rangle \label{main_proof}
\end{align}

To bound the first term in the RHS of Equation \eqref{main_proof}, we use Theorem \ref{thm_strongbabe1} with the choice of $\alpha$ as in the Theorem statement and for the error term $\varepsilon'$:
\begin{align}
\nonumber
   &f_\eta( \vX^{\eta, \alpha})  \\
   &\overset{\eqref{pqpot_objective_new}}{=} F_\eta(\vX^{\eta, \alpha}, \alpha)  -   P( \vX^{\eta, \alpha}, \alpha) \leq   f_\eta(\vX^\eta)  -   P( \vX^{\eta, \alpha}, \alpha)  + \varepsilon'  \leq f_\eta(\vX^\eta)+ \varepsilon' \notag \\
   \therefore  & \langle \mathbf{C}, \vX^{\eta, \alpha} - \vX^\eta \rangle \leq \eta\big( R(\vX^\eta) - R(\vX^{\eta, \alpha }) \big) +\varepsilon' \leq \eta R(\vX^\eta)  +\varepsilon'\leq  \eta U_R  +\varepsilon' \label{main_proof_1st_term}
\end{align}

Furthermore, by the formulations of the RPOT and the POT problem and Lemma \ref{lma_bound_regularizer}, we have the following bound:
\begin{align}
\nonumber
      \langle \mathbf{C}, \vX^\eta - \vX^* \rangle &= f_\eta( \vX^\eta ) - f_\eta(\vX^*) +  \eta\big(   R(\vX^*)  - R(\vX^\eta) \big) \\
      &\leq \eta R(\vX^*) \leq   \eta U_R,  \label{main_proof_2nd_term}
\end{align}

Applying Equations \eqref{main_proof_1st_term}, \eqref{main_proof_2nd_term}, \eqref{main_proof_4th_term}, and Hölder's inequality, \eqref{main_proof} becomes:

\begin{align}
\nonumber
     &\langle \mathbf{C}, \bar{\vX} \rangle - \langle \mathbf{C}, \vX^* \rangle  \\
     &=   
     \langle \mathbf{C}, \vX^{\eta, \alpha} - \vX^\eta \rangle  +   
     \langle \mathbf{C}, \vX^\eta - \vX^* \rangle + 
     \langle \mathbf{C}, \bar{\vX}-     \vX^k \rangle + 
     \langle \mathbf{C}, \vX^k - \vX^{\eta, \alpha}\rangle \notag
     \\
     &\notag \leq 2\eta U_R + \varepsilon' +  \|\vC\|_\infty\|\bar{\vX} - \vX^k\|_1 + \|\vC\|_\infty\|\vX^k - \vX^{\eta, \alpha}\|_1 \\
     &\leq 2\eta U_R + \varepsilon' + \sqrt{2}n (\eta\mu)^{-\frac{1}{2}}\varepsilon_1^{\frac{1}{2}}\|\vC\|_\infty + \|\vC\|_\infty\|\bar{\vX} - \vX^k\|_1  \label{main_proof_1st_step}
\end{align}
Then, it remains to bound the final term $\|\bar{\vX} - \vX^k\|_1$. From Lemma \ref{slack_bound}, we already obtained a bound for $\|\bar{\vX} - \vX^k\|_1$ as $23\delta$. In the following Lemma, we bound $\delta$.

\begin{lemma}\footnote{Proof for this Lemma in Appendix \ref{proof_5.4}}
    \label{slack_true_bound}
    We can establish the following bound on $\delta$, the constraint violation of $\vX^k$:
    \begin{align*}
        &\sum^n_{i=1}\max\{0,  (\vX^k\ones_n)_i-\vr_i\} + \max\{0, ((\vX^k)^\top\ones_n)_i - \vc_i\} \\&\leq \sqrt{\frac{2n}{\alpha}P(\vX^{\eta, \alpha},\alpha)} + 2\|\vX^k - \vX^{\eta, \alpha}\|_1
    \end{align*}
\end{lemma}

Hence, from the result of Lemma \ref{slack_bound}, Lemma \ref{slack_true_bound}, and Equation \eqref{main_proof_4th_term}, we obtain the following bound for the final term:
\begin{align}
    \|\bar{\vX} - \vX^k\|_1 &\leq 23\sqrt{ \frac{2n}{\alpha} P(\vX^{\eta,\alpha}, \alpha)} + 46\left\| \vX^k - \vX^{\eta,\alpha} \right\|_1 \notag\\
    &\leq 23\sqrt{ \frac{2n}{\alpha} P(\vX^{\eta,\alpha}, \alpha)} + 46\sqrt{2}n (\eta\mu)^{-\frac{1}{2}}\varepsilon_1^{\frac{1}{2}}
    \label{main_proof_3rd_term_1st_step}
\end{align}
Recall the result of Theorem \ref{thm_strongbabe1} and the value of $\alpha = \frac{(\sqrt{2n}+1)(s \|\vC\|_\infty  + \eta U_R)^2}{2\zeta^2 \varepsilon'}$:
\begin{align}
    \frac{n}{\alpha} &= \frac{2n\zeta^2 \varepsilon'}{(s \|\vC\|_\infty  + \eta U_R)^2(\sqrt{2n}+1)} \notag\\
    &\leq \frac{2n\zeta^2 \varepsilon'}{(s \|\vC\|_\infty  + \eta U_R)^2\sqrt{2n}} = \frac{\sqrt{2n}\,\zeta^2 \varepsilon'}{(s \|\vC\|_\infty  + \eta U_R)^2}
    \label{main_proof_3rd_term_2nd_step}
\end{align}

Moreover, by Theorem~\ref{thm_strongbabe1}, we can derive a bound on $P(\vX^{\eta,\alpha}, \alpha)$ where: 
$$F_\eta(\vX^{\eta,\alpha}, \alpha) \leq f_\eta(\vX^\eta) = \textbf{RPOT}_\eta \text{ (the first direction of \eqref{equiv_champ_full})}$$ and $$f_\eta(\vX^{\eta,\alpha}) \geq v(\tau_\alpha) \geq v(0) - \varepsilon' = \textbf{RPOT}_\eta - \varepsilon' \\ \text{ (from \eqref{eq:value_dom} and \eqref{eq:value_drop})}$$ , we obtain:
\begin{align}
\nonumber
    P(\vX^{\eta, \alpha}, \alpha) &\overset{\eqref{pqpot_objective_new}}{=} F_\eta(\vX^{\eta, \alpha}, \alpha) - f_\eta(\vX^{\eta, \alpha}) \\
    &\leq \textbf{RPOT}_\eta - (\textbf{RPOT}_\eta - \varepsilon') = \varepsilon'. \label{main_proof_3rd_term_3rd_step}
\end{align}
Hence, applying \eqref{main_proof_3rd_term_2nd_step} and \eqref{main_proof_3rd_term_3rd_step} back to \eqref{main_proof_3rd_term_1st_step}, we reach:
\begin{align}
    \|\bar{\vX} - \vX^k\|_1 &\leq 23\sqrt{ \frac{2n}{\alpha} P(\vX^{\eta,\alpha}, \alpha)} + 46\sqrt{2}n (\eta\mu)^{-\frac{1}{2}}\varepsilon_1^{\frac{1}{2}}\notag\\
    &\leq 23\sqrt{\frac{2\sqrt{2n}\,\zeta^2 \varepsilon'}{(s\|\vC\|_\infty + \eta U_R)^2}\cdot \varepsilon'} + 46\sqrt{2}n (\eta\mu)^{-\frac{1}{2}}\varepsilon_1^{\frac{1}{2}} \notag \\
    &= 23\cdot 2^{3/4}\,\frac{n^{1/4}\,\zeta\,\varepsilon'}{s\|\vC\|_\infty + \eta U_R} + 46\sqrt{2}n (\eta\mu)^{-\frac{1}{2}}\varepsilon_1^{\frac{1}{2}} \notag\\
    &= W_1 + 46\sqrt{2}n (\eta\mu)^{-\frac{1}{2}}\varepsilon_1^{\frac{1}{2}}
    \label{main_proof_3rd_term_4th_step}
\end{align}
where
\begin{equation*}
    W_1 := 23\cdot 2^{3/4}\,\frac{n^{1/4}\,\zeta\,\varepsilon'}{s\|\vC\|_\infty + \eta U_R}.
\end{equation*}
Applying \eqref{main_proof_3rd_term_4th_step} back to \eqref{main_proof_1st_step}, we arrive at the final bound
\begin{align}
     &\langle \mathbf{C}, \bar{\vX} \rangle - \langle \mathbf{C}, \vX^* \rangle  \leq 2\eta U_R + \varepsilon' +  47\sqrt{2}n (\eta\mu)^{-\frac{1}{2}}\varepsilon_1^{\frac{1}{2}}\|\vC\|_\infty + \|\vC\|_\infty W_1
     \label{main_proof_2nd_step}
\end{align}
To obtain the objective gap in Theorem \ref{thm_final} from \eqref{main_proof_2nd_step}, we proceed to derive the values of $\varepsilon'$ and $\varepsilon_1$ with respect to $\varepsilon$ by bounding each of the four terms on right-hand side by $\frac{\varepsilon}{4}$. 

Balancing $\|\vC\|_\infty W_1 \leq \frac{\varepsilon}{4}$ gives
\begin{equation*}
    \|\vC\|_\infty \cdot 23 \cdot 2^{3/4}\,\frac{n^{1/4}\,\zeta\,\varepsilon'}{s\|\vC\|_\infty + \eta U_R} \leq \frac{\varepsilon}{4}
    \quad\Longleftrightarrow\quad
    \varepsilon' \leq \frac{(s\|\vC\|_\infty + \eta U_R)\,\varepsilon}{23 \cdot 2^{11/4}\,n^{1/4}\,\zeta\,\|\vC\|_\infty}.
\end{equation*}
Combined with the constraint $\varepsilon' \leq \varepsilon/4$ from the second term of \eqref{main_proof_2nd_step}, we hence derive the value of each quantity:
\begin{align*}
&\eta = \frac{\varepsilon}{8U_R} \Rightarrow s\|\vC\|_\infty + \eta U_R = s\|\vC\|_\infty + \frac{\varepsilon}{8}, \;\varepsilon_1 = \frac{\varepsilon^2(\eta\mu)}{32\cdot47^2n^2\|\vC\|^2_\infty} \\
&\varepsilon' = \min\left\{\frac{\varepsilon}{4},\, \frac{(s \|\vC\|_\infty  + \eta U_R)\,\varepsilon}{23 \cdot 2^{11/4}\,n^{1/4}\,\zeta\,\|\vC\|_\infty}\right\} = \Theta\!\left(\frac{\varepsilon}{n^{1/4}}\right).
\end{align*}

Moreover, we can derive the upper bound for $\|\vX^0 - \vX^{\eta, \alpha}\|^2_2$ using the constraint $\ones^\top\vX\ones = s$:
\begin{equation}
    \|\vX^0 - \vX^{\eta, \alpha}\|_2^2 \leq (\|\vX^0\|_2+\|\vX^{\eta, \alpha}\|_2)^2 \leq (\|\vX^0\|_1+\|\vX^{\eta, \alpha}\|_1)^2 \leq 4s^2
    \label{final_bound}
\end{equation}
Recall that $\kappa = \frac{\beta}{\eta\mu}$ and applying the newly derived values and \eqref{final_bound}, we arrive at the final complexity of $k$ for which Algorithm \ref{alg:PNAG} achieves $\varepsilon$-approximate solution:
\begin{align}
    k &\geq \sqrt{\kappa}\left[\ln\!\left(\frac{(\eta\mu)\|\vX^0 - \vX^{\eta, \alpha}\|_2^2}{2\varepsilon_1}\right) + \ln\!\left(\frac{23}{3}\right) \right] \notag\\
    &=  \sqrt{\frac{\eta L + 4n\alpha}{\eta\mu}}\left[\ln\!\left(\frac{64\cdot 47^{2}\,n^{2}\,s^{2}\,\|\vC\|_{\infty}^{2}}{\varepsilon^{2}}\right) + \ln\!\left(\frac{23}{3}\right)\right] \notag\\
    &=\mathcal{O}\!\left(\sqrt{\frac{n\alpha}{\eta}}\,\ln\!\left(\frac{n}{\varepsilon}\right)\right)
    = \mathcal{O}\!\left(\frac{n^{7/8}}{\varepsilon}\,\ln\!\left(\frac{n}{\varepsilon}\right)\right)
\end{align}\qed

\section{Experiments}

We empirically benchmark PNAG-POT against RPOT baselines on three real-world tasks: color transfer \cite{poulicolortransfer}, point cloud registration \cite{shen-point-cloud,bai2023partialtransportpointcloudregistration}, and domain adaptation \cite{redkodomainadaptaion}. We consider both quadratic and Elastic-Net regularization. Our PNAG-POT is self-implemented and uses line search initialized with $t_0=\frac{1}{\beta}$, while the Quadratic-regularized POT (QPOT) baseline and Elastic-Net-regularized POT (ENPOT) are solved using CVXPY with the first-order SCS solver \cite{scs}, ensuring methodological consistency. We also include APDAGD from \cite{our_pot} as an additional baseline.

PNAG-POT is compared to CVXPY in terms of final cost, sparsity, iteration count, and wall-clock runtime. Even though APDAGD targets entropic regularization, we also include it as one of the baselines. Across all experiments, we fix the transport mass $s = 0.9\min\{\|\vr\|_1, \|\vc\|_1\}$, transferring $90\%$ of the source mass. The regularization weight $\eta$ is fine-tuned empirically, while $\alpha$ is calibrated to the scale of the final objective for effective penalization and numerical stability, with Theorem~\ref{thm_strongbabe1} providing a conservative sufficient theoretical guarantee for $\varepsilon$-approximation. Sparsity is measured by thresholding entries below $10^{-8}$. In our experiments, we initialized $\vX$ as random.

The task formulations, cost matrices, transported mass, and parameter-selection strategy are shared between the quadratic and Elastic-Net experiments. We therefore describe each task once below and subsequently report the results under the two regularizers.

\subsection{Experimental Tasks}

\subsubsection{Color Transfer}

In the color transfer task, we transport the color distribution of a source image to emulate that of a reference image, harmonizing visual tone while preserving spatial structure. We formulate the RPOT problem over smoothed and normalized $n=1024$-bin histograms of RGB values from each image. The cost matrix is defined by the squared Euclidean distance between the centers of the source and target histogram bins.

\subsubsection{Point Cloud Registration}

Point clouds are sets of 3D points captured by sensors such as LiDAR \cite{pomerleau2015review} or multi-view stereo \cite{ylimaki2015fa}, often used to represent real-world scenes in applications like autonomous driving \cite{geiger2012we} and robotics \cite{pomerleau2015review}. Partial point cloud registration seeks a transformation that aligns two point clouds of unequal size. Partial Optimal Transport (POT) offers a suitable framework for this task.

In our experiments, we represent the point clouds as uniform discrete distributions, each containing \(n=500\) points, with the cost matrix defined by squared Euclidean distances. The goal is to transport mass from the source (blue) to the target (red).

\subsubsection{Domain Adaptation}

Domain adaptation addresses the problem of adapting data distributions between a labeled source domain and an unlabeled target domain, enabling transfer learning across different domains. In this experiment, we use the moons dataset from scikit-learn \cite{pedregosa2011scikit}, with both the source and target represented as uniform discrete distributions of $n=300$ samples. The cost matrix is defined by the normalized pairwise Euclidean distances between these distributions.

We evaluate the performance of the algorithms based on transport sparsity, final cost, and adaptation accuracy by training a Support Vector Machine (SVM) on the transported domain and evaluate on the target domain.

\subsection{Quadratic Regularization}

Table~\ref{tab:qpot_results} summarizes the quantitative results under quadratic regularization. APDAGD runs on entropic regularized POT is also included as one of the baselines.

\begin{table}[h!]
\centering
\caption{Comparison of the methods under quadratic regularization.}
\label{tab:qpot_results}
\small
\setlength{\tabcolsep}{3pt}

\begin{tabular}{@{}llccccc@{}}
\toprule
Task & Method 
& \makecell{Final\\Cost} 
& \makecell{Raw Cost\\$\inner{\vC}{\vX}$}
& Sparsity 
& Iterations 
& \makecell{Time\\(s)} \\
\midrule

\multirow{3}{*}{Color Transfer}
& CVXPY
& $0.0278$
& $0.0278$
& $0.9848$
& $50{,}000$
& $6007.00$ \\

& APDAGD
& $0.0440$
& $0.0440$
& $0.9805$
& $10{,}256$
& $741.07$ \\

& PNAG-POT
& \textbf{$0.0272$}
& \textbf{$0.0272$}
& \textbf{$0.9958$}
& \textbf{$1{,}800$}
& \textbf{$173.55$} \\

\midrule

\multirow{3}{*}{\makecell{Point Cloud\\ Registration}}
& CVXPY
& $0.0146$
& $0.0146$
& $0.4351$
& $11{,}650$
& $820.14$ \\

& APDAGD
& $0.0116$
& $0.0116$
& $0.8424$
& $9{,}000$
& $570.95$ \\

& PNAG-POT
& $0.0084$
& $0.0084$
& $0.9229$
& $2{,}940$
& $65.62$ \\

\midrule

\multirow{3}{*}{Domain Adaptation}
& CVXPY
& $0.0357$
& $0.0357$
& $0.4532$
& $8{,}500$
& $210.09$ \\

& APDAGD
& $0.0366$
& $0.0366$
& $0.6937$
& $11{,}152$
& $244.34$ \\

& PNAG-POT
& $0.0356$
& $0.0356$
& $0.9180$
& $1{,}420$
& $9.32$ \\

\bottomrule
\end{tabular}
\end{table}

\subsubsection{Color Transfer}

\begin{figure}[H]
    \centering
    \includegraphics[width=0.6\linewidth]{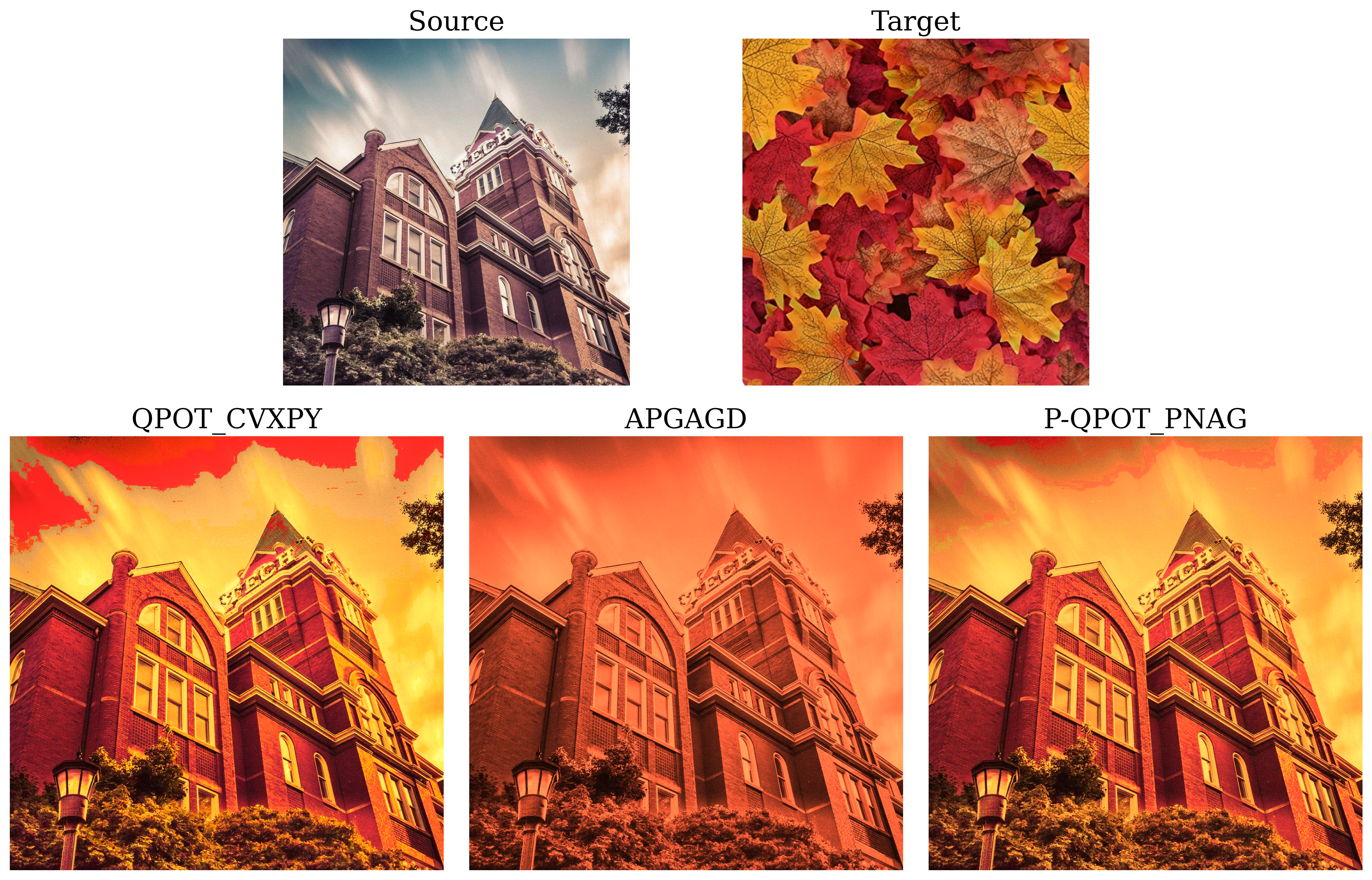}
    \caption{Original source and target images and the source image with colors transferred using the compared POT methods.}
    \label{fig:CF}
\end{figure}

We set the regularization parameter to $\eta = 10^{-3}$. For PNAG-POT, we use a penalty parameter of $\alpha=5500$. Figure~\ref{fig:CF} shows the original source and target images together with the transferred results.

\begin{figure}[H]
    \centering
    \includegraphics[width=\linewidth]{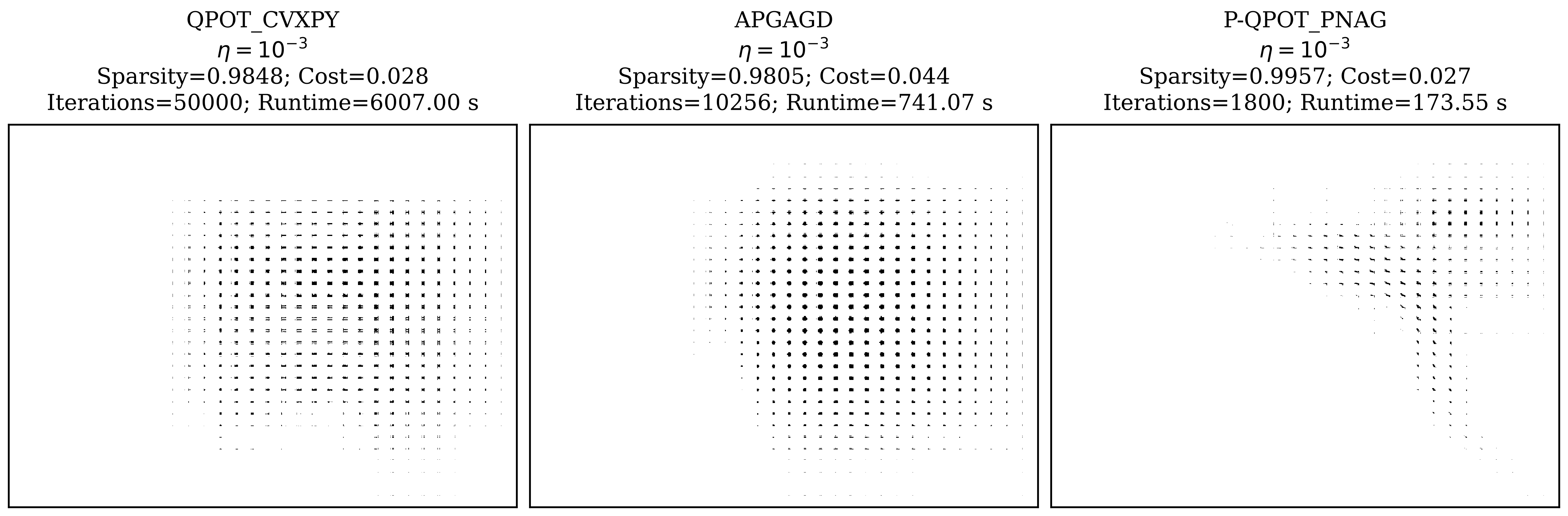}
    \caption{Heatmaps of transport plans for color transfer obtained by (a) CVXPY, (b) APDAGD, and (c) PNAG-POT.}
    \label{fig:CF_result}
\end{figure}

Figure~\ref{fig:CF_result} shows the transport plans produced by CVXPY, APDAGD, and PNAG-POT. As summarized in Table~\ref{tab:qpot_results}, PNAG-POT obtains the lowest final cost of $0.0272$ and the highest sparsity of $0.9958$, while requiring only $1{,}800$ iterations compared with $50{,}000$ for CVXPY. Compared with APDAGD, PNAG-POT also achieves a substantially lower final cost while producing a slightly sparser transport plan.

\subsubsection{Point Cloud Registration}

\begin{figure}[H]
    \centering
    \includegraphics[width=\linewidth]{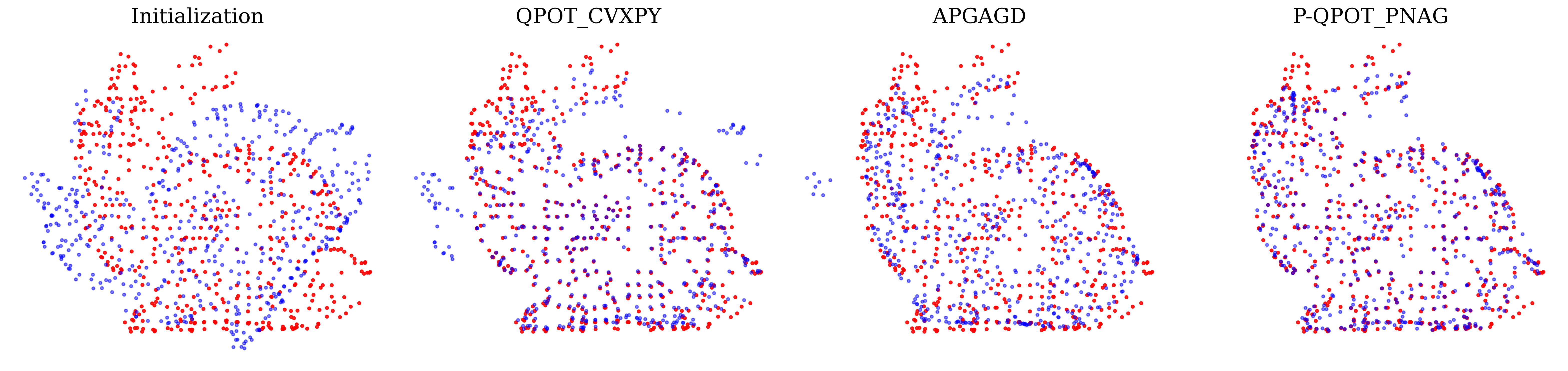}
    \caption{(a) Original point clouds, (b) transported point clouds using CVXPY, and (c) transported point clouds using PNAG-POT. The point clouds are visualized in 2D instead of 3D.}
    \label{fig:PC}
\end{figure}

We set $\eta=10^{-3}$ and also use a penalty parameter of $\alpha=5500$ for PNAG-POT. Figure~\ref{fig:PC} illustrates the original point clouds and the resulting registration. Compared with CVXPY, PNAG-POT produces a more compact and coherent alignment on the rabbit dataset, particularly in the left half of the shape.

\begin{figure}[H]
    \centering
    \includegraphics[width=\linewidth]{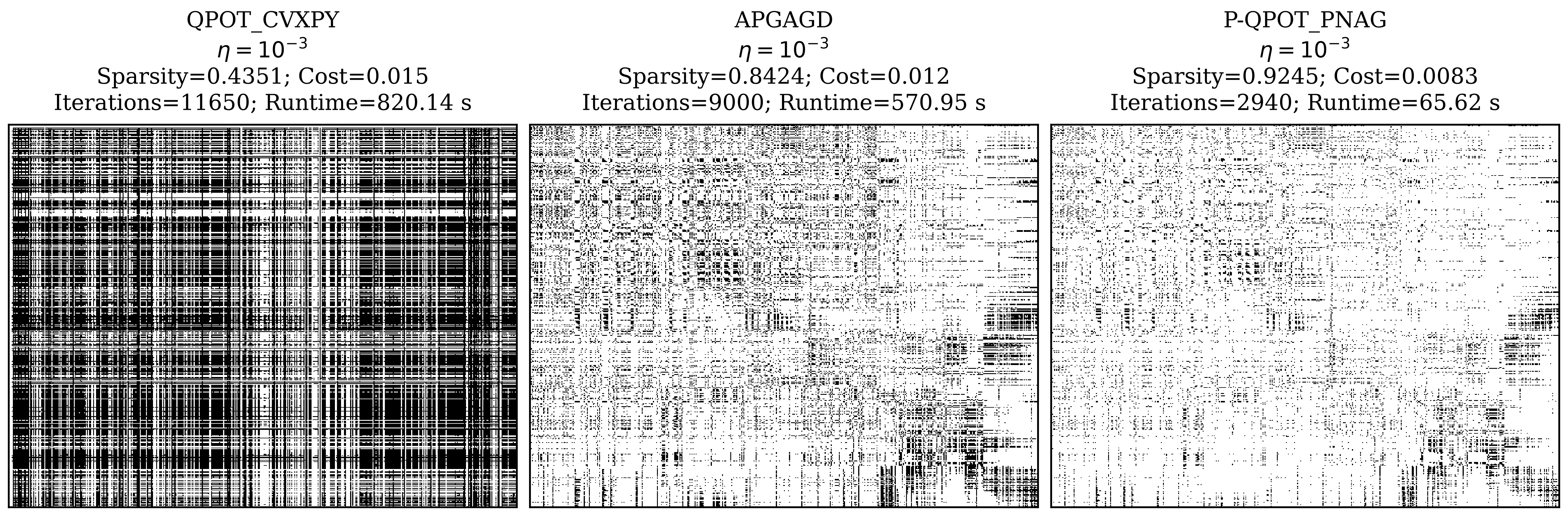}
    \caption{Heatmaps of transport plans for point cloud registration obtained by (a) CVXPY, (b) APDAGD, and (c) PNAG-POT.}
    \label{fig:PC_result}
\end{figure}

Figure~\ref{fig:PC_result} compares the corresponding transport plans. PNAG-POT achieves the lowest final cost of $0.0084$ and the highest sparsity of $0.9229$. It requires $2{,}940$ iterations compared with $11{,}650$ for CVXPY, while reducing the runtime from $820.14$ to $65.62$ seconds. Compared with APDAGD, PNAG-POT also yields a lower-cost and sparser transport plan.

\subsubsection{Domain Adaptation}

\begin{figure}[H]
    \centering
    \includegraphics[width=0.9\linewidth]{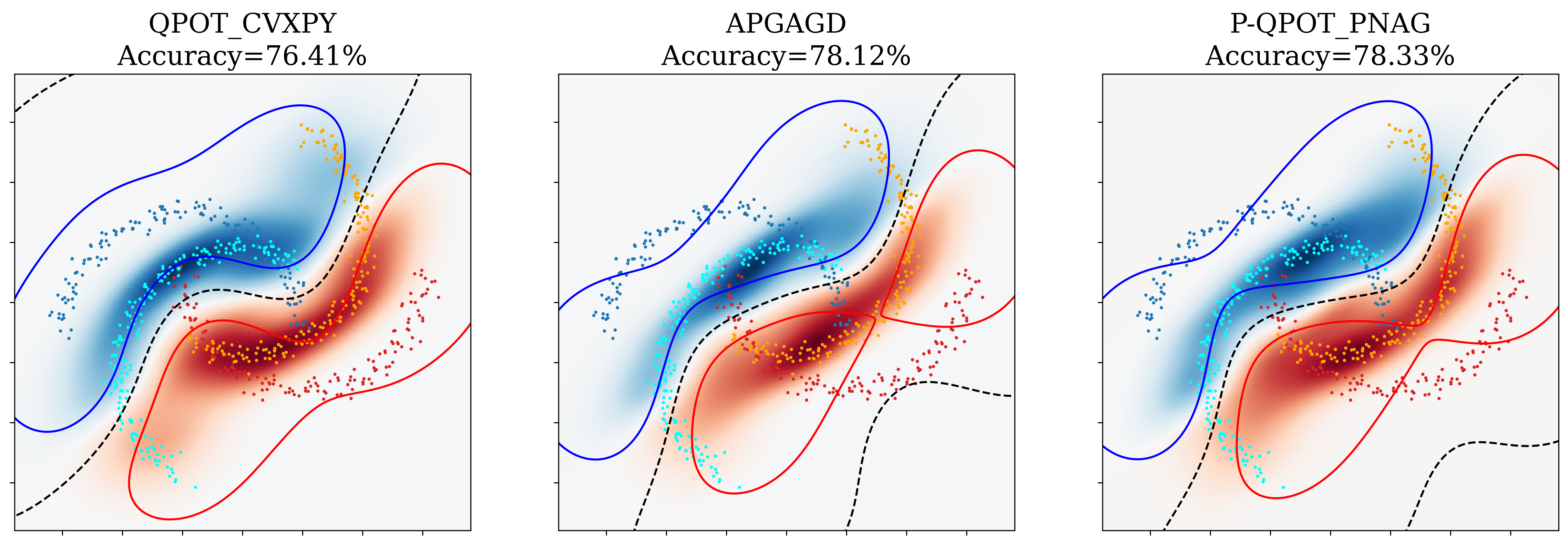}
    \caption{Decision boundaries of the SVM on the transported data obtained by (a) CVXPY, (b) APDAGD, and (c) PNAG-POT.}
    \label{fig:DA}
\end{figure}

For domain adaptation, we set $\eta=10^{-3}$ and use penalty parameter $\alpha=55000$. Figure~\ref{fig:DA} illustrates the domain adaptation results on the moons dataset, including the SVM decision boundaries derived from the adapted data.

\begin{figure}[H]
    \centering
    \includegraphics[width=\linewidth]{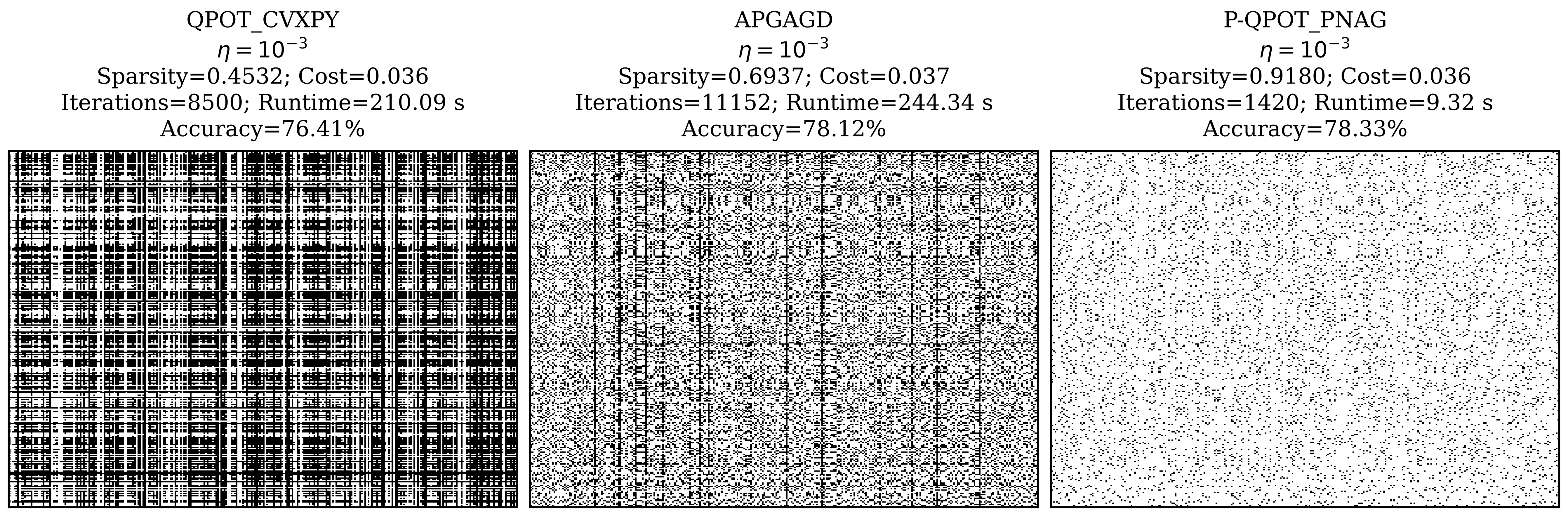}
    \caption{Heatmaps of transport plans for domain adaptation obtained by (a) CVXPY, (b) APDAGD, and (c) PNAG-POT.}
    \label{fig:DA_result}
\end{figure}

Figure~\ref{fig:DA_result} shows the corresponding transport plans. PNAG-POT reaches the same final cost of $0.036$ as CVXPY while increasing sparsity from $0.4532$ to $0.9180$ and reducing the iteration count from $8{,}500$ to $1{,}420$. The runtime is reduced from $210.09$ to $9.32$ seconds. PNAG-POT also achieves an adaptation accuracy of $78.33\%$, compared with $76.41\%$ for CVXPY and $78.12\%$ for APDAGD, indicating that the substantially sparser transport plan does not come at the expense of adaptation performance.

\subsection{Elastic-Net Regularization}

We further evaluate PNAG-POT with Elastic-Net regularization to verify that the proposed optimization framework is not restricted to quadratic regularization. We follow the same task formulations, transport mass, cost matrices, evaluation metrics, and parameter-selection strategy described above. Therefore, we do not repeat the experimental setup for each task. Recall the formula of the Element-wise Elastic Net regularizers $
R(X)\;=\;\frac{\eta_2}{2}\|\vX\|_2^2\;+\;\|\vW_1\odot\vX\|_{1}$, we set regularization weight $\eta_2=10^{-3}$ and weight matrix $\vW_1$ where $W_{1, ij} = 10^{-3} + 10^{-4}1_{i=j}$ across all experiments. This introduces a simple heterogeneous regularization setting for numerical evaluation. Finally, we retain the same penalty parameter $\alpha$ used in the corresponding quadratic-regularization experiments.

Table~\ref{tab:enpot_results} summarizes the results obtained by CVXPY and PNAG-POT under Elastic-Net regularization. Across the three tasks, PNAG-POT obtains solutions with comparable or lower final cost while requiring fewer iterations and producing sparser transport plans. The corresponding behavior is consistent with that observed under quadratic regularization.

\begin{comment}
\begin{table}[h!]
\centering
\caption{Comparison between CVXPY and PNAG-POT under Elastic-Net regularization.}
\label{tab:enpot_results}
\small
\setlength{\tabcolsep}{3pt}

\begin{tabular}{@{}llccccc@{}}
\toprule
Task & Method 
& \makecell{Final\\Cost} 
& \makecell{Raw Cost\\$\inner{\vC}{\vX}$}
& Sparsity 
& Iterations 
& \makecell{Time\\(s)} \\
\midrule

\multirow{2}{*}{Color Transfer}
& CVXPY
& $0.0287$
& $0.0278$
& $0.9847$
& $50{,}000$
& $7607.85$ \\

& PNAG-POT
& $0.0281$
& $0.0272$
& $0.9959$
& $1{,}800$
& $620.22$ \\

\midrule

\multirow{2}{*}{\makecell{Point Cloud\\Registration}}
& CVXPY
& $0.0157$
& $0.0148$
& $0.4218$
& $11{,}675$
& $489.58$ \\

& PNAG-POT
& $0.0093$
& $0.0083$
& $0.9281$
& $2{,}940$
& $204.49$ \\

\midrule

\multirow{2}{*}{Domain Adaptation}
& CVXPY
& $0.0366$
& $0.0357$
& $0.4644$
& $8{,}025$
& $70.64$ \\

& PNAG-POT
& $0.0365$
& $0.0356$
& $0.9181$
& $1{,}420$
& $8.46$ \\

\bottomrule
\end{tabular}
\end{table}
\end{comment}

\begin{table}[h!]
\centering
\caption{Comparison between CVXPY and PNAG-POT under Weighted Elastic-Net regularization.}
\label{tab:enpot_results}
\small
\setlength{\tabcolsep}{3pt}

\begin{tabular}{@{}llccccc@{}}
\toprule
Task & Method 
& \makecell{Final\\Cost} 
& \makecell{Raw Cost\\$\inner{\vC}{\vX}$}
& Sparsity 
& Iterations 
& \makecell{Time\\(s)} \\
\midrule

\multirow{2}{*}{Color Transfer}
& CVXPY
& $0.0450$
& $0.0441$
& $0.9920$
& $12{,}425$
& $8340.01$ \\

& PNAG-POT
& $0.0277$
& $0.0268$
& $0.9961$
& $2{,}800$
& $713.50$ \\

\midrule

\multirow{2}{*}{\makecell{Point Cloud\\Registration}}
& CVXPY
& $0.0155$
& $0.0146$
& $0.4398$
& $11{,}800$
& $1642.86$ \\

& PNAG-POT
& $0.0092$
& $0.0083$
& $0.9327$
& $3{,}140$
& $572.59$ \\

\midrule

\multirow{2}{*}{Domain Adaptation}
& CVXPY
& $0.0376$
& $0.0367$
& $0.4551$
& $8{,}475$
& $346.37$ \\

& PNAG-POT
& $0.0370$
& $0.0361$
& $0.9086$
& $1{,}515$
& $26.57$ \\

\bottomrule
\end{tabular}
\end{table}

\begin{figure}[H]
    \centering
    \includegraphics[width=0.9\linewidth]{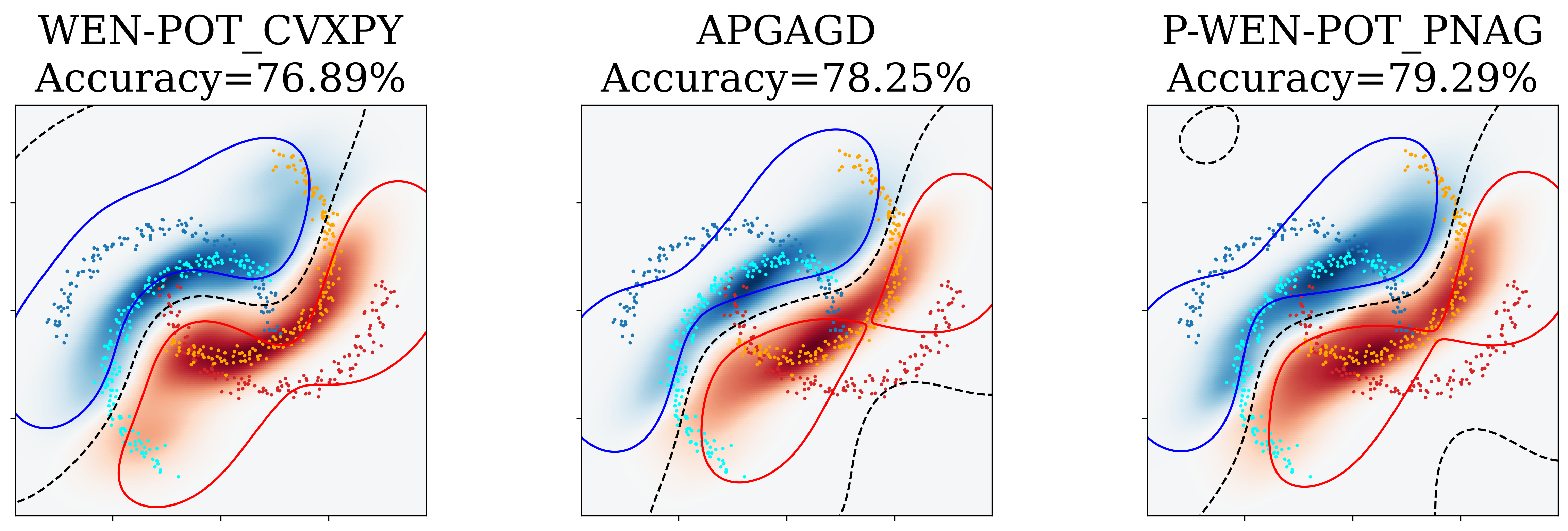}
    \caption{Domain adaptation results under Elastic-Net regularization obtained by (a) CVXPY, (b) APDAGD, and (c) PNAG-POT.}
    \label{fig:DA_ENPOT}
\end{figure}

Figure~\ref{fig:DA_ENPOT} shows the corresponding domain adaptation results under Weighted Elastic-Net regularization. For domain adaptation, PNAG-POT also achieves higher accuracy of $79.29\%$, compared with $76.89\%$ for CVXPY and $78.25\%$ for APDAGD while maintaining cost and requiring considerably fewer iterations.

The corresponding Weighted Elastic-Net results for color transfer and point cloud registration, together with the transport-plan visualizations for all three tasks, are reported in the Appendix~\ref{app:elastic_net_results}.
\section{Conclusions}
We develop a dedicated first-order solver for Regularized Partial Optimal Transport (RPOT) with smooth and strongly convex regularizers that promotes sparse transport plan, including quadratic and elastic-net regularization. By reformulating RPOT using an exterior quadratic penalty framework on the marginal constraints, we relax the strict constraints while preserving smoothness and strong convexity, allowing our accelerated gradient method, PNAG-POT, to efficiently obtain an approximate solution that can then be rounded to a feasible plan whose cost is within $\varepsilon$ of the original optimum. We further benchmarked our methodology on three machine learning applications: color transfer, domain adaptation, and point cloud registration and over two choices of regularizers. Our P-RPOT formulation, along with our proposed PNAG-POT algorithm, consistently converges to a lower-cost and sparser transport plan with higher efficiency. We believe that our method could be further extended to other settings or used to improve computational methods for other variants of Optimal Transport, such as Unbalanced Optimal Transport \cite{gem-uot,le2021robust} or Constrained Optimization problems \cite{Lan2016ConditionalGS,wai-defw,nguyen2024stochasticconstraineddecentralizedoptimization,tang2025a-constrained-ot} in general. Improving the computational methods for these problems could potentially improve the performance in many Machine Learning applications, such as Few-shot Learning \cite{bach-fsar,Liu_2025_Graph}, Generative Modeling \cite{vayer2019sliced,balaji2020robust}, or Time Series Modeling \cite{pmlr-v108-janati20a-temporal-ot,timeseries1}, Healthcare \cite{uot-healthcare}, as well as scalable adoption in large-scale systems \cite{xgboost,error_hpc1,error_hpc2,nguyen2021understandingscalabilityhyperledgerfabric}.
\begin{comment}
\begin{acknowledgements}
Acknowledgements to sponsoring agencies and individuals should be placed here.
\end{acknowledgements}
\end{comment}

\newpage
\appendix  


\section{Appendix A: Guarantees for the quadratic exterior penalty \cite{penalty_mazal1}}
\label{app:truemper}

We restate the results of \cite{penalty_mazal1} used in the proof of Theorem~\ref{thm_strongbabe1}. \cite{penalty_mazal1} formulation is a maximization problem $f$ over $x\in E^n$ with penalty $-r\sum_i[\min(0,g_i(x))]^2$. In our case, we restate the formulation in our minimization form of this paper, with $\mathcal{S}$ kept as a set constraint. The detailed proofs are in \cite{penalty_mazal1}.

We write the $m = 2n$ marginal constraints as $g_i(\vX) \ge 0$, where for $i = 1,\dots,n$
\begin{equation*}
g_i(\vX) = r_i - (\vX\ones_n)_i,
\qquad
g_{n+i}(\vX) = c_i - (\vX^\top\ones_n)_i .
\end{equation*}
Given the penalty~\eqref{penalty_func1} $P(\vX,\alpha) = \alpha\sum_{i=1}^{m}\min\{0, g_i(\vX)\}^2$ and the marginal constraints $g_i(\vX)$. The Lagrangian is
\begin{equation*}
L(\vX,\vu) = f_\eta(\vX) - \sum_{i=1}^{m} u_i\, g_i(\vX),
\qquad \vX\in\mathcal{S},\ \vu\in\RR^m_+ ,
\end{equation*}
and $(\vX^0,\vu^0)$ is a saddle point of $L$ if
\begin{equation*}
L(\vX^0,\vu) \le L(\vX^0,\vu^0) \le L(\vX,\vu^0)
\qquad \forall\,\vX\in\mathcal{S},\ \vu\in\RR^m_+ .
\end{equation*}
Under Slater's condition (Definition~\ref{def:slater-pot}), such a saddle point exists with $\vX^0 = \vX^\eta$~\cite[\S5.4.2]{boyd_convex_2004}. The penalty threshold $\alpha_0$ is required to satisfy~\cite[Eq.~(7)]{penalty_mazal1}
\begin{equation}
\alpha_0 > \max_i u_i^0 .
\label{eq:app-threshold}
\end{equation}

\begin{proposition}[{\cite[Eqs.~(8)--(9)]{penalty_mazal1}}]
\label{prop:app-threshold}
Let $\bar{\vX}\in\mathcal{S}$ satisfy $g_i(\bar{\vX}) > 0$ for all $i$, and let $z$ be a lower bound on $f_\eta(\vX^0)$. If
\begin{equation*}
\alpha_0 > \frac{f_\eta(\bar{\vX}) - z}{\min_i g_i(\bar{\vX})},
\end{equation*}
then $\alpha_0 > \|\vu^0\|_1 \ge \max_i u_i^0$. In particular, $\alpha_0$ satisfies~\eqref{eq:app-threshold}.
\end{proposition}

\begin{theorem}[{\cite[Theorem~1]{penalty_mazal1}}]
\label{thm:app-truemper}
Assume that $L$ has a saddle point $(\vX^0,\vu^0)$ and that $\alpha_0$ satisfies~\eqref{eq:app-threshold}. For any $\tau > 0$, set
\begin{equation*}
h = \frac{\sqrt{m}+1}{2},
\qquad
\alpha = \frac{\alpha_0\, h}{\tau}.
\end{equation*}
Then every minimizer $\vX^\star$ of $F_\eta(\cdot,\alpha)$ over $\mathcal{S}$ satisfies $g_i(\vX^\star) \ge -\tau$ for all $i = 1,\dots,m$, that is,
\begin{equation*}
\vX^\star\ones_n - \tau\ones_n \le \vr,
\qquad
(\vX^\star)^\top\ones_n - \tau\ones_n \le \vc.
\end{equation*}
In particular, $\vX^\star$ is $\tau$-convergent in the sense of Definition~\ref{eps-convergence}.
\end{theorem}

Although Theorem~\ref{thm:app-truemper} concerns the quadratic penalty, its threshold~\eqref{eq:app-threshold} and Proposition~\ref{prop:app-threshold} are derived from Truemper's analysis of linear penalties and remain valid in the quadratic case.

\section{Appendix B: Algorithm: Projection of a matrix on to $\Ss$}
\label{app:alg_projection}

\begin{algorithm}[H]
\caption{Euclidean projection of a matrix onto $\Ss$ via the probability simplex}
\label{alg:simplex_projection}
\begin{minipage}{\linewidth}
\textbf{Require:} $\mathbf{Y} \in \mathbb{R}^{n \times n}$, scalar $s > 0$
\vspace{1mm}
\begin{enumerate}
  \item Flatten $\mathbf{Y}$ to a 1D vector $\mathbf{y} \in \mathbb{R}^{n^2}$
  \item Normalize: $\tilde{\mathbf{y}} \leftarrow \mathbf{y} / s$
  \item Sort $\tilde{\mathbf{y}}$ into $\mathbf{u}$ such that $u_1 \geq u_2 \geq \cdots \geq u_{n^2}$
  \item Find $\rho = \max \left\{1 \leq j \leq n^2 : u_j + \frac{1}{j} \left(1 - \sum_{i=1}^j u_i \right) > 0 \right\}$
  \item Define $\lambda = \frac{1}{\rho} \left( 1 - \sum_{i=1}^\rho u_i \right)$
  \item Compute $\mathbf{x}'$ such that $x'_i = \max \{\tilde{y}_i + \lambda, 0\}, \quad i = 1, \ldots, n^2$
  \item Rescale: $\mathbf{x} \leftarrow s \cdot \mathbf{x}'$
  \item Reshape $\mathbf{x}$ back to matrix form with shape $(n, n)$
\end{enumerate}
\vspace{1mm}
\textbf{Return:} Matrix $\mathbf{X}$ satisfying $X_{ij} \geq 0$ and $\sum_{i,j} X_{ij} = s$
\end{minipage}
\end{algorithm}

\begin{comment}
\section*{Appendix D. The Proximal Gradient Descent Algorithm}

\begin{algorithm}[H]
\caption{Proximal Gradient Descent}
\label{alg:proximal-gradient-descent}
\begin{minipage}{\linewidth}
\textbf{Require:} Initial point $\mathbf{X}_0$, step size $t > 0$
\vspace{1mm}

\begin{enumerate}
  \item \textbf{For} $k = 1, 2, \dots$ \textbf{do}
  \begin{enumerate}
    \item Compute gradient step: 
    $$
    \mathbf{y}^{(k)} = \mathbf{x}^{(k-1)} - t \nabla g(\mathbf{x}^{(k-1)})
    $$
    \item Apply projection step: 
    $$
    \mathbf{x}^{(k)} = \text{prox}_{t}(\mathbf{y}^{(k)})
    $$
  \end{enumerate}
\end{enumerate}
\vspace{1mm}
\textbf{Return:} Final iterate $\mathbf{x}^{(k)}$
\end{minipage}
\end{algorithm} 
\end{comment}

\section{Appendix C: Guarantees for ROUND-POT \cite[Theorem 6]{our_pot}}
\label{app:theorem6ourpot}
We restate the guarantee of ROUND-POT (Algorithm~\ref{alg:RoundPOT}) used in the
proof of Lemma~\ref{slack_bound}, with $\vA$, $\vb$ and
$\vx = (\operatorname{vec}(\vX)^\top, \vp^\top, \vq^\top)^\top \in \RR^{n^2+2n}$
as defined there. The proof is in \cite{our_pot}.
\begin{theorem}[{\cite[Theorem 6]{our_pot}}]
\label{thm:roundpot_guarantee}
If $\vx \geq 0$ and $\|\vA \vx - \vb\|_1 \leq \delta$ for some $\delta \geq 0$, then
Algorithm~\ref{alg:RoundPOT} returns in $\mathcal{O}(n^2)$ time a vector
$\bar{\vx} = (\operatorname{vec}(\bar{\vX})^\top, \bar{\vp}^\top, \bar{\vq}^\top)^\top \geq 0$
with
\begin{equation*}
\vA \bar{\vx} = \vb
\qquad \text{and} \qquad
\|\vx - \bar{\vx}\|_1 \leq 23\,\delta .
\end{equation*}
In particular $\bar{\vX} \in \mathcal{U}(\vr,\vc,s)$ and
$\|\vX - \bar{\vX}\|_1 \leq 23\,\delta$.
\end{theorem}

\newpage

\section{Appendix D: Proof of Lemma \ref{slack_true_bound}}
\label{proof_5.4}
{\it Proof } 
We can derive the bound for the constraint violations as follows
\begin{align*}
&\sum^n_{i=1}\left[\max\{0, (\vX^k\ones_n)_i-\vr_i \} + \max\{0, ((\vX^k)^\top\ones_n)_i-\vc_i\}\right] \\ 
&= \sum_{i=1}^{n} \left[
- \min\left(0,\; \vr_i - (\vX^k \mathbf{1}_n)_i\right)
- \min\left(0,\; \vc_i - ((\vX^k)^\top \mathbf{1}_n)_i\right)
\right] \\
&= \sum_{i=1}^{n} \left[
- \min\left(0,\; \vr_i - (\vX^{\eta,\alpha} \mathbf{1}_n)_i\right)
- \min\left(0,\; \vc_i - ((\vX^{\eta,\alpha})^\top \mathbf{1}_n)_i\right)
\right] \\
&\quad + \sum_{i=1}^{n} \left[
\left( - \min\left(0,\; \vr_i - (\vX^k \mathbf{1}_n)_i\right) 
+ \min\left(0,\; \vr_i - (\vX^{\eta,\alpha} \mathbf{1}_n)_i\right) \right) \right. \\
&\quad\quad + \left.
\left( - \min\left(0,\; \vc_i - ((\vX^k)^\top \mathbf{1}_n)_i\right) 
+ \min\left(0,\; \vc_i - ((\vX^{\eta,\alpha})^\top \mathbf{1}_n)_i\right) \right)
\right] \\
&\leq  \sum_{i=1}^{n} \left[
- \min\left(0,\; \vr_i - (\vX^{\eta,\alpha} \mathbf{1}_n)_i\right)
- \min\left(0,\; \vc_i - ((\vX^{\eta,\alpha})^\top \mathbf{1}_n)_i\right)
\right] \\
&\quad + \sum_{i=1}^{n} \left[
\left| \min\left(0,\; \vr_i - (\vX^k \mathbf{1}_n)_i\right) 
- \min\left(0,\; \vr_i - (\vX^{\eta,\alpha} \mathbf{1}_n)_i\right) \right| \right. \\
&\quad\quad + \left.
\left| \min\left(0,\; \vc_i - ((\vX^k)^\top \mathbf{1}_n)_i\right) 
- \min\left(0,\; \vc_i - ((\vX^{\eta,\alpha})^\top \mathbf{1}_n)_i\right) \right|
\right]
\end{align*}
 
Moreover, we know that 
\begin{equation}
\label{ineq}
\left| \min\{0,u\} - \min\{0,v\} \right| =
\begin{cases}
|u - v| & \text{if } u \leq 0,\; v \leq 0 \\
0       & \text{if } u \geq 0,\; v \geq 0 \\
|v|     & \text{if } u > 0 \geq v         \\
|u|     & \text{if } v > 0 \geq u
\end{cases}
\leq |u - v| \qquad \text{for all } u,v \in \mathbb{R}.
\end{equation}

Now using the just proven result from \eqref{ineq}, we have:
\begin{align}
\nonumber
    &\sum^n_{i=1}\max\{0, (\vX^k\ones_n)_i-\vr_i\} + \max\{0, ((\vX^k)^\top\ones_n)_i-\vc_i\} \\ 
    &\leq   \sum^n_{i=1}\left[ - \min(0, \vr_i - (\mathbf{X}^{\eta,\alpha} \ones_n)_i) - \min(0, \vc_i - ( (\mathbf{X}^{\eta,\alpha})^\top \ones_n)_i)\right]\\
    \nonumber
    &\quad +   \sum^n_{i=1}\left[ | (\mathbf{X}^{k} \ones_n)_i-  (\mathbf{X}^{\eta,\alpha} \ones_n)_i |     + | ((\mathbf{X}^{k})^\top \ones_n)_i-((\mathbf{X}^{\eta,\alpha})^\top \ones_n)_i  |  \right]\\
    &\leq   \sum^n_{i=1}\left[ - \min(0, \vr_i - (\mathbf{X}^{\eta,\alpha} \ones_n)_i) - \min(0, \vc_i - ( (\mathbf{X}^{\eta,\alpha})^\top \ones_n)_i)\right] + 2 \|\vX^k -\vX^{\eta,\alpha} \|_1 . \label{approx_baby1}
\end{align}

Now applying Cauchy-Schwarz,  we get for any $\vX^{\eta, \alpha}$:
\begin{align}
   &\bigg[ \sum^n_{i=1}\left( - \min(0, \vr_i - (\mathbf{X}^{\eta,\alpha}\ones_n)_i) - \min(0, \vc_i - ((\mathbf{X}^{\eta,\alpha})^\top\ones_n)_i)\right) \bigg]^2 \notag\\
   & \leq 2n \sum_{i=1}^{n} \left(\min\left(0,\; \vr_i - (\mathbf{X}^{\eta,\alpha}\ones_n)  \right)^2 + \min\left(0,\; \vc_i - ((\mathbf{X}^{\eta,\alpha})^\top \mathbf{1}_n)_i \right)^2 \right) \notag\\
   & \leq \frac{2n}{\alpha} \alpha \sum_{i=1}^{n} \left(\min\left(0,\; \vr_i - (\mathbf{X}^{\eta,\alpha} \mathbf{1}_n)_i \right)^2 + \min\left(0,\; \vc_i - ((\mathbf{X}^{\eta,\alpha})^\top \mathbf{1}_n)_i \right)^2 \right) \notag\\ 
   &\leq 2n \cdot  \frac{P(\mathbf{X}^{\eta,\alpha}, \alpha)}{\alpha} \label{approx_baby2}
\end{align}

Combining inequalities \eqref{approx_baby1} and \eqref{approx_baby2}, we obtain:
\begin{align}
&\sum^n_{i=1}\max\{0, (\vX^k\ones_n)_i-\vr_i\} + \max\{0, ((\vX^k)^\top\ones_n)_i-\vc_i\} \\ 
&\leq \sum_{i=1}^n \left[ - \min\left(0,\, r_i - (\mathbf{X}^{\eta,\alpha} \mathbf{1}_n)_i \right) - \min\left(0,\, c_i - ((\mathbf{X}^{\eta,\alpha})^\top \mathbf{1}_n)_i \right) \right] 
+ 2 \left\| \vX^k - \vX^{\eta,\alpha} \right\|_1 \notag \\
&\leq  
\sqrt{ \frac{2n}{\alpha} P(\vX^{\eta,\alpha}, \alpha)} + 2 \left\| \vX^k - \vX^{\eta,\alpha} \right\|_1 \label{eq:final_bound}
\end{align}
which concludes the proof of the Lemma. 
\qed

\section{Appendix E: Additional Results for Elastic-Net Regularization}
\label{app:elastic_net_results}
This section provides additional qualitative results for the Elastic-Net experiments reported in the main manuscript. We include the task-specific results for color transfer and point cloud registration, together with the corresponding transport-plan visualizations for all three tasks.
\subsection{Color Transfer}
\begin{figure}[H]
    \centering
    \includegraphics[width=0.6\linewidth]{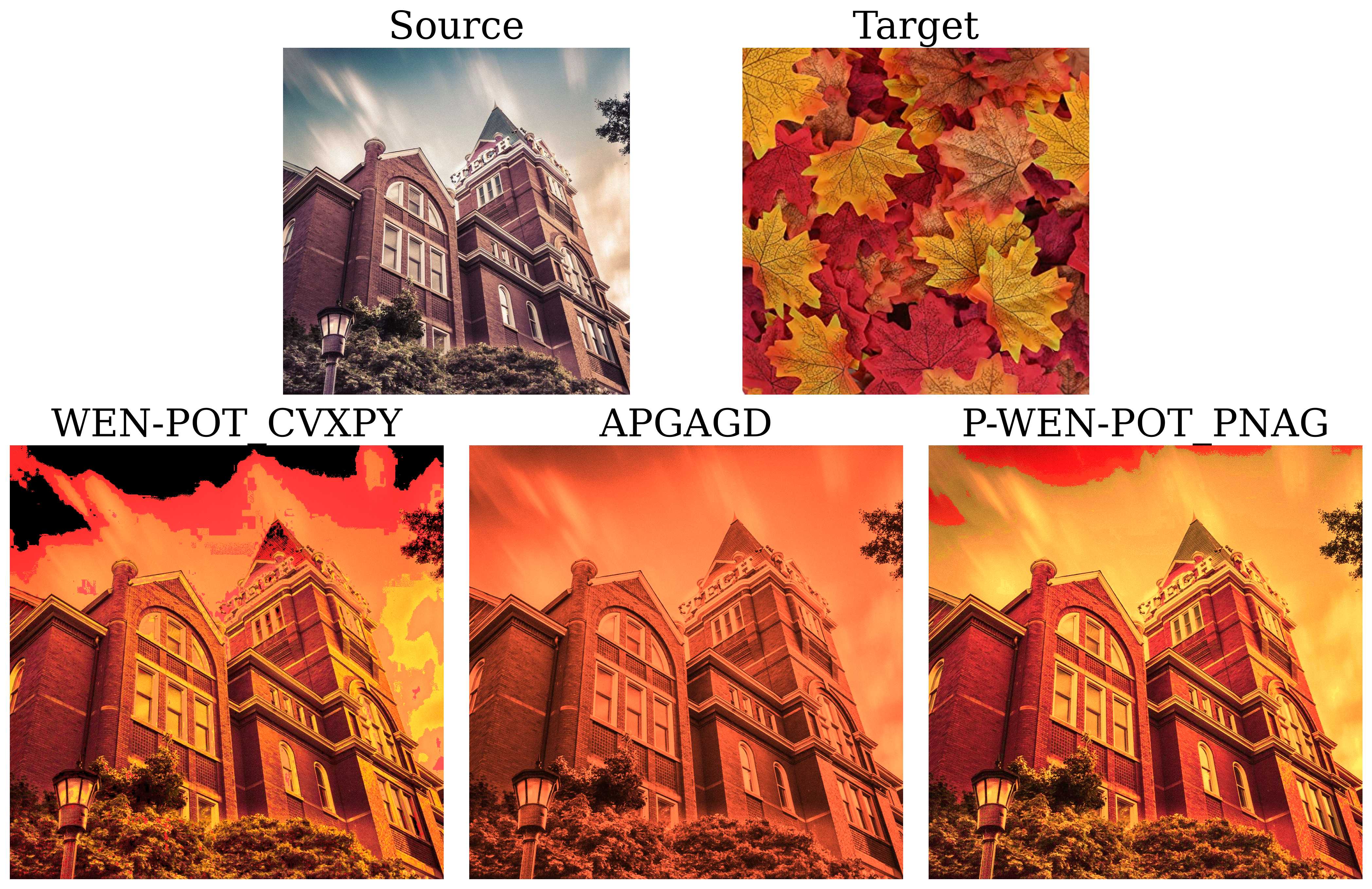}
    \caption{Color-transfer results under Elastic-Net regularization. The figure shows the source and target images together with the transferred results obtained by the compared POT methods.}
    \label{fig:CF_ENPOT_appendix}
\end{figure}

\begin{figure}[H]
    \centering
    \includegraphics[width=\linewidth]{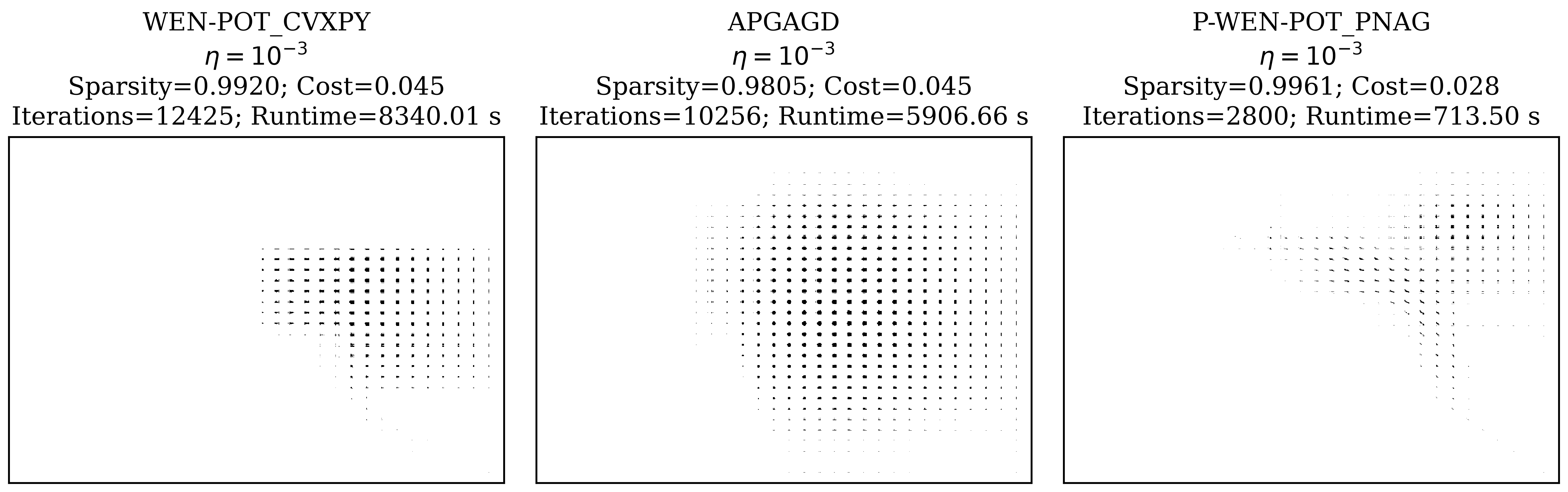}
    \caption{Transport-plan heatmaps for color transfer under Elastic-Net regularization.}
    \label{fig:CF_ENPOT_heatmap_appendix}
\end{figure}

Figure~\ref{fig:CF_ENPOT_appendix} shows the color-transfer results obtained under Weighted Elastic-Net regularization. PNAG-POT produces a transferred image comparable to the CVXPY solution while requiring substantially fewer iterations. Figure~\ref{fig:CF_ENPOT_heatmap_appendix} further shows the corresponding transport plans, where PNAG-POT yields a visibly sparser solution.
\subsection{Point Cloud Registration}
\begin{figure}[h!]
    \centering
    \includegraphics[width=\linewidth]{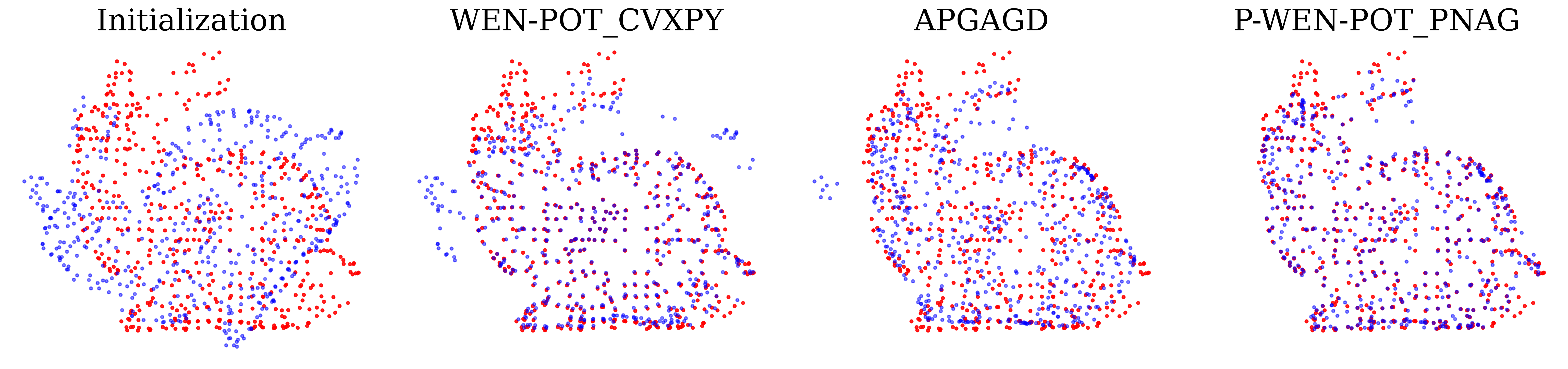}
    \caption{Point-cloud registration results under Elastic-Net regularization.}
    \label{fig:PC_ENPOT_appendix}
\end{figure}

\begin{figure}[h!]
    \centering
    \includegraphics[width=\linewidth]{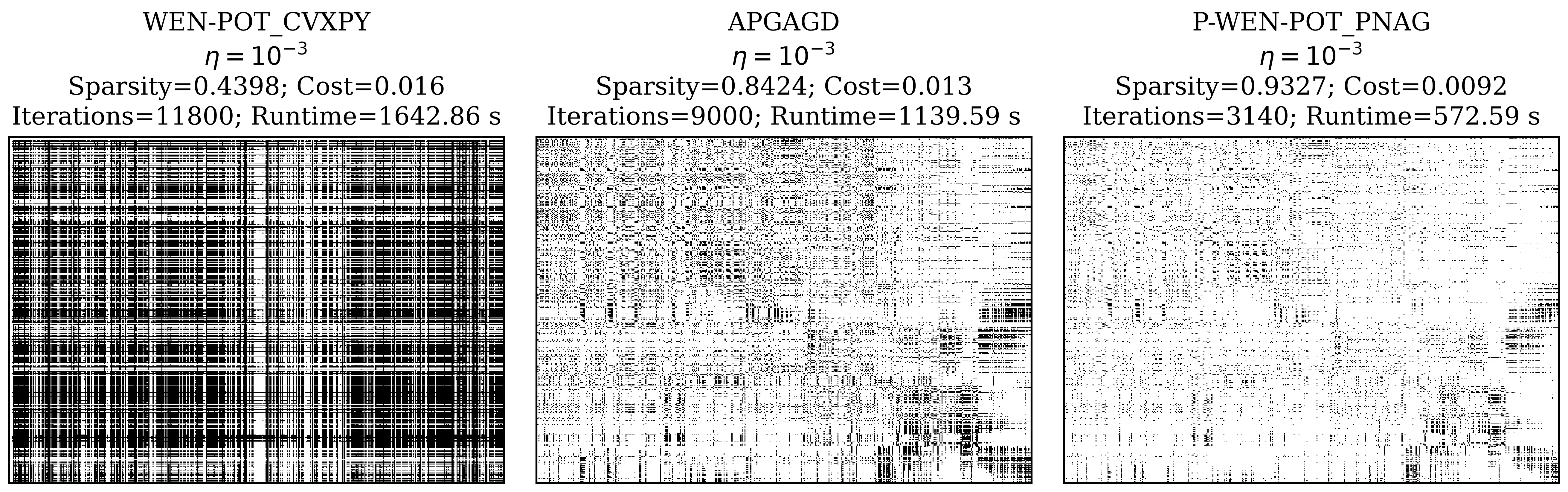}
    \caption{Transport-plan heatmaps for point cloud registration under Elastic-Net regularization.}
    \label{fig:PC_ENPOT_heatmap_appendix}
\end{figure}

Figure~\ref{fig:PC_ENPOT_appendix} shows the point-cloud registration results under Elastic-Net regularization. PNAG-POT produces a compact alignment of the transported point cloud while preserving the overall geometric structure. The corresponding transport plans are shown in Figure~\ref{fig:PC_ENPOT_heatmap_appendix}, where PNAG-POT produces a substantially sparser transport plan than CVXPY.
\subsection{Domain Adaptation}
\begin{figure}[h!]
    \centering
    \includegraphics[width=0.9\linewidth]{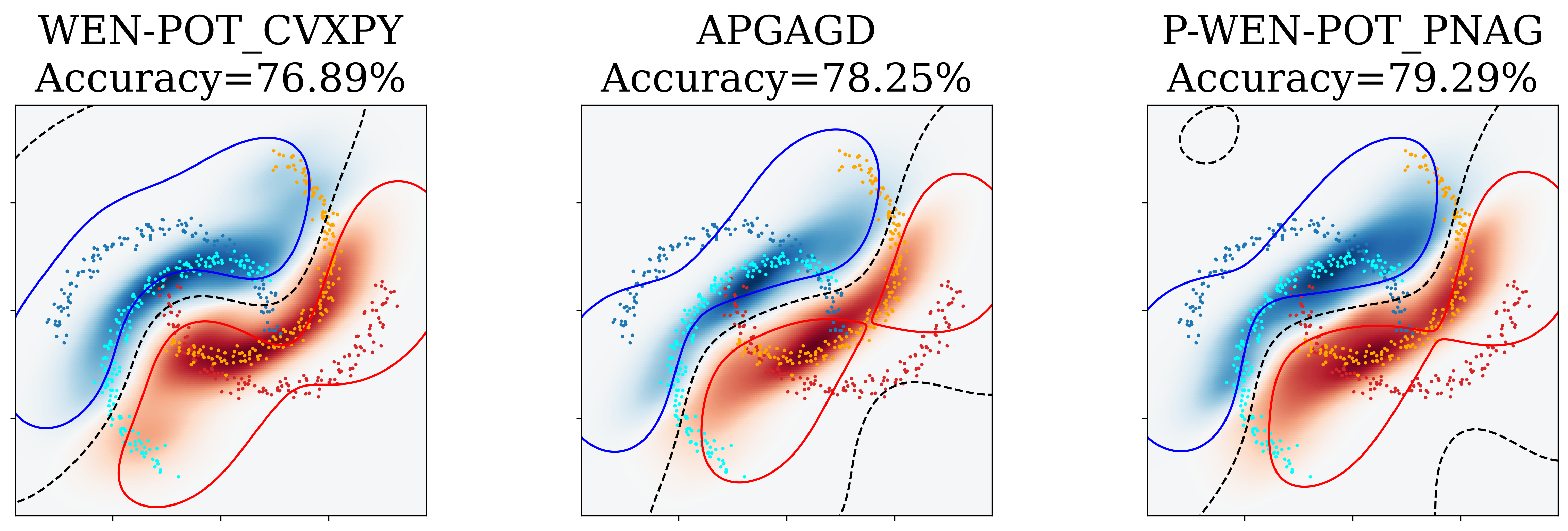}
    \caption{Domain-adaptation results under Elastic-Net regularization.}
    \label{fig:DA_ENPOT_appendix}
\end{figure}

\begin{figure}[h!]
    \centering
    \includegraphics[width=\linewidth]{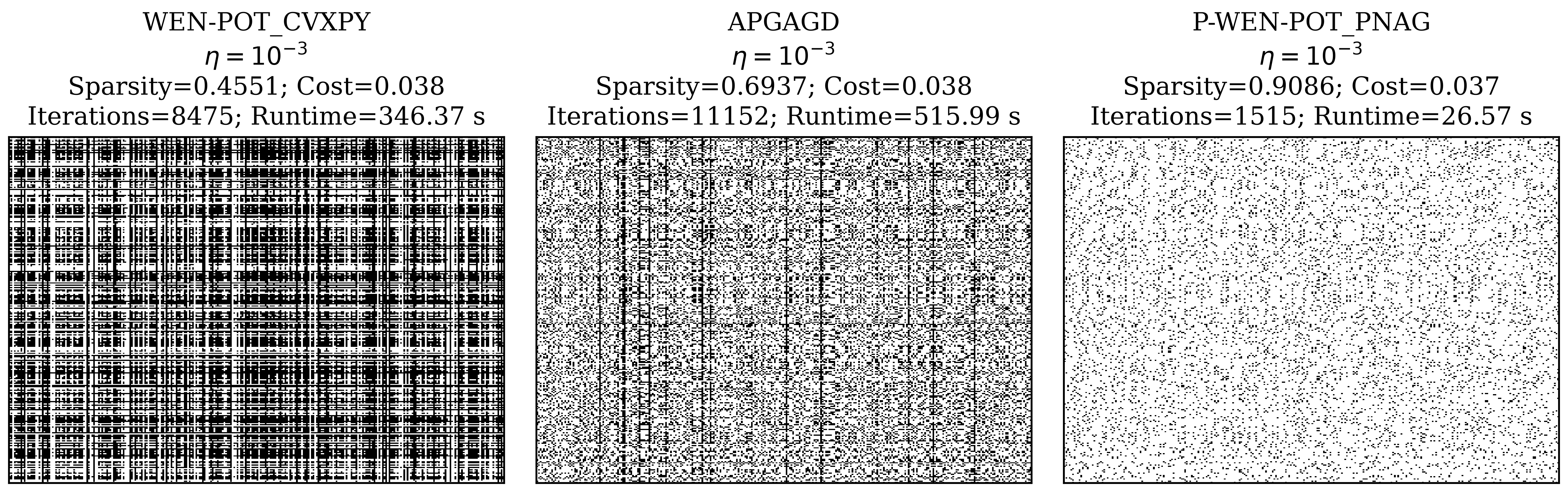}
    \caption{Transport-plan heatmaps for domain adaptation under Elastic-Net regularization.}
    \label{fig:DA_ENPOT_heatmap_appendix}
\end{figure}

Figure~\ref{fig:DA_ENPOT_appendix} shows the domain-adaptation results under Weighted Elastic-Net regularization. PNAG-POT maintains the adaptation performance while producing a solution comparable to the other methods. Figure~\ref{fig:DA_ENPOT_heatmap_appendix} shows the corresponding transport plans, where PNAG-POT yields a considerably sparser solution.

\pagebreak
\newpage
\bibliographystyle{spmpsci}
\bibliography{refs}


\end{document}